\documentclass{article}
\usepackage{spconf,amsmath,graphicx,nccmath}

\usepackage{graphicx}
\usepackage{subcaption}
\usepackage{booktabs}
\usepackage{mathtools}
\usepackage[export]{adjustbox}
\usepackage{enumitem}
\usepackage{xcolor}
\usepackage{array}
\usepackage{amsfonts}
\usepackage{float}
\usepackage{graphbox}
\usepackage{makecell}
\usepackage{multirow}
\usepackage{tabularx}
\usepackage[hidelinks]{hyperref}
\newcolumntype{Y}{>{\centering\arraybackslash}X}

\AtBeginEnvironment{flalign}{%
  \setlength{\abovedisplayskip}{5pt} % Change 5pt to your desired value
  \setlength{\belowdisplayskip}{5pt} % Change 5pt to your desired value
}

\AtBeginEnvironment{equation}{%
  \setlength{\abovedisplayskip}{5pt} % Change 5pt to your desired value
  \setlength{\belowdisplayskip}{5pt} % Change 5pt to your desired value
}

\DeclareTextFontCommand{\hyphtexttt}{\ttfamily\hyphenchar\font=45\relax}

\definecolor{lorange}{RGB}{255, 243, 224}
\definecolor{borderblue}{RGB}{68, 114, 196}
\definecolor{bluegray}{RGB}{56,74,99}
\definecolor{lightblue}{RGB}{71,118,204}
\definecolor{img_green}{RGB}{2,254,3}
\definecolor{img_blue}{RGB}{74,118,198}
\colorlet{RED}{red}
\colorlet{GREEN}{green}

\colorlet{BLUE}{blue}

\newcommand\Tstrut{\rule{0pt}{2.6ex}}       % "top" strut
\newcommand\Bstrut{\rule[-0.9ex]{0pt}{0pt}} % "bottom" strut
\title{SE3D: A Framework for Saliency Method Evaluation in 3D Imaging}  % Volumetric?
\name{Mariusz Wiśniewski, Loris Giulivi, Giacomo Boracchi
\thanks{Giacomo Boracchi was partially supported by the ``PNRR-PE-AI FAIR'' and the ``AI for Sustainable Port-city logistics (PNNR Grant P2022FLLPY)'' projects funded by the NextGeneration EU program.}}
\address{
 DEIB, Politecnico di Milano, Italy\\
}
\begin{document}
\ninept
\maketitle
\begin{abstract}
%---------------------------------------------------

% IMPORTANT!!!
% ABSTRACT LIMITED TO 200 WORDS. DO NOT CHANGE!

For more than a decade, deep learning models have been dominating in various 2D imaging tasks. Their application is now extending to 3D imaging, with 3D Convolutional Neural Networks (3D CNNs) being able to process LIDAR, MRI, and CT scans, with significant implications for fields such as autonomous driving and medical imaging.
In these critical settings, explaining the model's decisions is fundamental. Despite recent advances in Explainable Artificial Intelligence, however, little effort has been devoted to explaining 3D CNNs, and many works explain these models via inadequate extensions of 2D saliency methods. 

A fundamental limitation to the development of 3D saliency methods is the lack of a benchmark to quantitatively assess these on 3D data. To address this issue, we propose SE3D: a framework for \textbf{S}aliency method \textbf{E}valuation in \textbf{3D} imaging.
We propose modifications to ShapeNet, ScanNet, and BraTS datasets, and evaluation metrics to assess saliency methods for 3D CNNs.
We evaluate both state-of-the-art saliency methods designed for 3D data and extensions of popular 2D saliency methods to 3D. Our experiments show that 3D saliency methods do not provide explanations of sufficient quality, and that there is margin for future improvements and safer applications of 3D CNNs in critical fields.

\end{abstract}
\begin{keywords}
Deep Learning, Saliency Maps, 3D Convolutions, Computer Vision
\end{keywords}
\vspace{-7pt}
\section{Introduction}
\vspace{-7pt}
\label{sec:introduction}
% Deep learning models are dominating the computer vision scene, and have been employed for a variety of tasks including image classification \cite{inception,resnet}, object detection~\cite{faster_rcnn,yolo}, and semantic segmentation~\cite{unet}. 
% Despite their popularity, due to their complexity and large number of parameters, these models are effectively black-boxes, which makes their outputs difficult to interpret.
% This problem is exacerbated when deep neural networks are applied in critical domains where trustworthiness is essential, such as medical imaging. 
% Indeed, amongst countless applications, deep learning models have been used to discern between cancerous and non-cancerous nuclei in histological images~\cite{colon_cancer_classification,breast_cancer_classification}, detect lesions in CT scans~\cite{DeepLesion}, and segment organs in X-ray images~\cite{xray_segmentation1, xray_segmentation2}.
Deep learning models dominate the computer vision landscape. Enabled by the ever-increasing computational power and data availability, deep models have been able to solve a growing array of tasks. Notably, this trend extends to the domain of 3D data, where 3D Convolutional Neural Networks (3D CNNs) have demonstrated impressive performance across diverse applications. For example, 3D CNNs have been used in medical imaging to detect lesions in CT scans~\cite{DeepLesion}, and in autonomous driving to perform object detection from voxelized LIDAR scans~\cite{AutonomousDrivingVoxelNet}.
%in the field of medical imaging, these models have been successfully employed for tasks such as classifying cancerous nuclei in histological images~\cite{colon_cancer_classification}, detecting lesions in CT scans~\cite{DeepLesion}, and segmenting organs in X-ray images~\cite{xray_segmentation1}.
%Deep learning models dominate the computer vision scene. Thanks to increased computational power and data availability, an ever increasing number of tasks can be solved with deep models. 
%This is the particular case of 3D data, where recently 3D Convolutional Neural Networks (CNNs) have achieved impressive performance on a plethora of tasks. For example, in medical imaging, these models have been applied to a variety of tasks, including classification of cancerous nuclei in histological images~\cite{colon_cancer_classification}, lesion detection in CT scans~\cite{DeepLesion}, and organ segmentation in X-ray images~\cite{xray_segmentation1}.
%In medical imaging, these models have been applied to a variety of tasks, including classification of cancerous nuclei in histological images~\cite{colon_cancer_classification}, lesion detection in CT scans~\cite{DeepLesion}, and organ segmentation in X-ray images~\cite{xray_segmentation1}.

\begin{figure}[t]
    \centering
    \renewcommand\tabcolsep{0pt}
    \renewcommand\theadfont{\large}
    \begin{adjustbox}{max width=0.55\textwidth}
        \hspace{0.09\textwidth}
        \begin{tabularx}{\textwidth}{p{20pt}ccc}
            ~ & \textbf{\thead{Saliency map \\ over input}} & \textbf{\thead{WSOL evaluation \\ over segmentation}} & \textbf{\thead{WSSS evaluation \\ over segmentation}} \\
            \rotatebox[origin=c]{90}{\large{2D saliency maps}} & \setlength{\fboxsep}{10pt}\setlength{\fboxrule}{0.0pt}\fbox{\includegraphics[width=100pt,align=c]{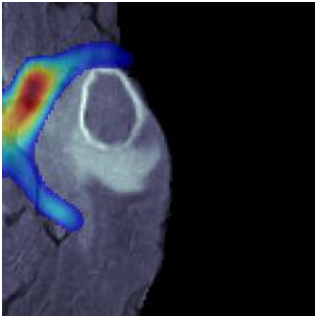}} & \setlength{\fboxsep}{10pt}\setlength{\fboxrule}{0.0pt}\fbox{\includegraphics[width=100pt,align=c]{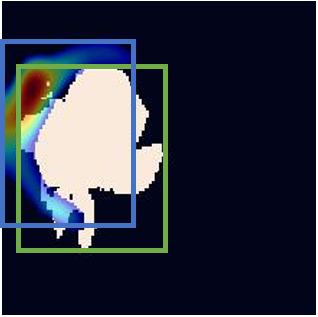}}& \setlength{\fboxsep}{10pt}\setlength{\fboxrule}{0.0pt}\fbox{\includegraphics[width=100pt,align=c]{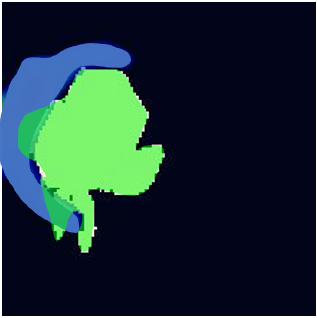}} \\
            \rotatebox[origin=c]{90}{\large{3D saliency maps}} & \setlength{\fboxsep}{10pt}\setlength{\fboxrule}{0.0pt}\fbox{\includegraphics[width=100pt,align=c]{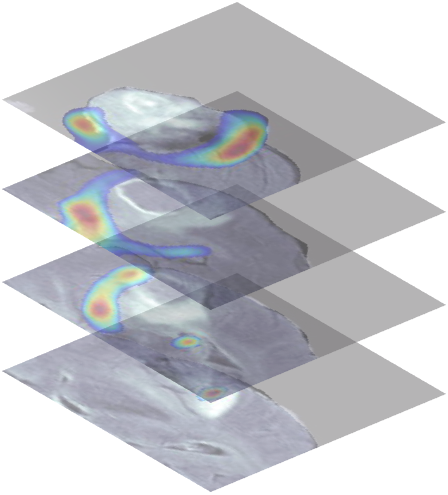}} & \setlength{\fboxsep}{10pt}\setlength{\fboxrule}{0.0pt}\fbox{\includegraphics[width=100pt,align=c]{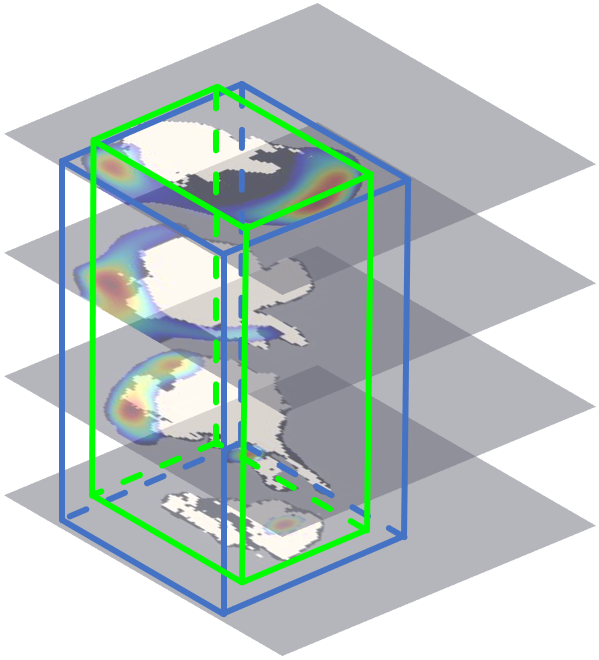}}& \setlength{\fboxsep}{10pt}\setlength{\fboxrule}{0.0pt}\fbox{\includegraphics[width=100pt,align=c]{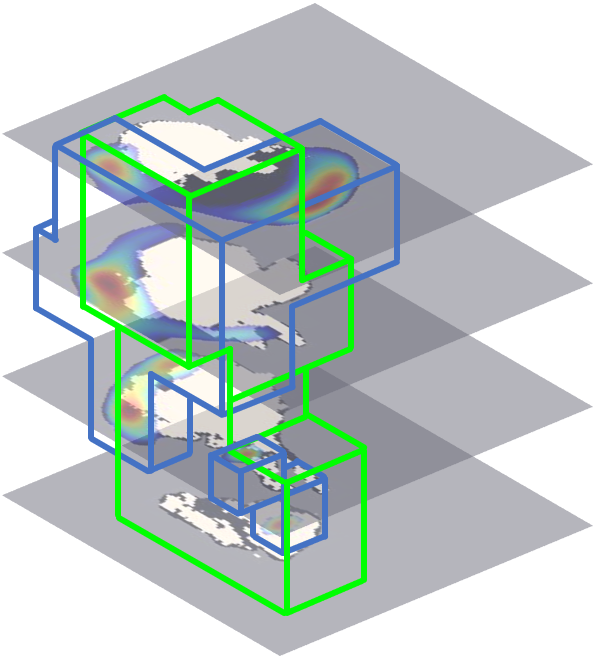}} \\
        \end{tabularx}
    \end{adjustbox}
    \vspace{-13pt}
    \caption{Example saliency map for a tumor classification model. We highlight the 2D/3D bounding boxes (center) and pixel/voxel-wise segmentations (right) of the {\color{img_green} ground truth tumor} and {\color{img_blue} saliency map}. The saliency map is a heatmap indicating the input regions that are most relevant to the model's output. For 2D data, evaluation is typically done by measuring WSOL and WSSS performance on localization and segmentation datasets (top). In this work, we propose a framework to extend the evaluation to 3D data and models (bottom).}
    \label{fig:teaser}
    \vspace{-8pt}
\end{figure}

Deep learning models, however, are hardly interpretable. To explain them, a plethora of methods have been proposed, with saliency methods (Section \ref{sec:background}) being the most popular.
Other than explaining network outputs, saliency methods are of particular interest in fields such as medical imaging, as they can be employed to easily obtain pixel-level annotations. Indeed, through Weakly Supervised Object Localization (WSOL) and Weakly Supervised Semantic Segmentation (WSSS) procedures (Figure \ref{fig:teaser}), saliency maps provide segmentation masks from models trained only on image-level supervision. 

Despite the surge of interest in 3D CNNs,
%\cite{3d_cnn_rt_object_recognition, 3d_cnn_human_action_recognition, 21d_cnn, 3d_csn}, 
little progress has been made towards developing 3D saliency methods.
%and solving WSOL and WSSS tasks on 3D images. 
Furthermore, while 2D saliency methods can be rigorously evaluated on the WSOL and WSSS tasks \cite{WSOLRight} (Figure \ref{fig:teaser}), there is no benchmark to evaluate the performance of saliency methods for 3D CNNs.
SE3D lifts this limitation and enables rigorous evaluation of saliency methods on 3D data. We propose a framework consisting of \textbf{performance metrics} and \textbf{datasets} built upon modified versions of the ShapeNet~\cite{shapenet}, ScanNet~\cite{scannet}, and BraTS~\cite{BraTS1, BraTS2, BraTS3} 3D datasets. These constitute a series of benchmarks for extensive evaluation of 3D CNN saliency methods.

% Our work lifts this limitation by proposing SE3D: a rigorous framework to assess saliency methods for 3D CNNs. SE3D consists of \textbf{performance metrics} and \textbf{datasets} built upon modified versions of the ShapeNet~\cite{shapenet}, ScanNet~\cite{scannet}, and BraTS~\cite{BraTS1, BraTS2, BraTS3} 3D datasets, constituting a series of benchmarks that enable extensive evaluation.

%modified versions of popular 3D medical imaging datasets \cite{BraTS1, BraTS2, BraTS3, KiTS21}
%, medical_decathlon}.

% \begin{itemize}
%     \item \textbf{evaluation metrics} for 3D saliency methods based on weakly supervised object localization and weakly supervised semantic segmentation (Figure \ref{fig:teaser}, \ref{fig:metrics}),
%     \item \textbf{a benchmark} built upon three 3D medical imaging datasets \cite{XXX}, for which we propose transformations to enable 3D saliency map evaluation.
    
%     %classification dataset} built upon the BraTS 2020 \cite{BraTS1, BraTS2, BraTS3} brain tumor segmentation dataset, including \textit{a pre-trained 3D CNN classification model} (Figure \ref{fig:dataset_preparation}).
% \end{itemize}

%\noindent To the best of our knowledge, we are the first to propose a benchmark to rigorously evaluate 3D saliency methods. 
In our experiments, we evaluate 3D-specific saliency methods (Saliency Tubes \cite{saliency_tubes}, Respond-CAM \cite{Respond-CAM}), and extensions of the most popular 2D saliency methods to 3D data (Grad-CAM \cite{Grad-CAM}, Grad-CAM++ \cite{Grad-CAM++}, and HiResCAM \cite{HiResCAM}).
%a technique designed specifically for 3D models, Saliency Tubes \cite{saliency_tubes}, and compare it against our extensions of 2D saliency methods such as Grad-CAM \cite{Grad-CAM}, Grad-CAM++ \cite{Grad-CAM++}, and HiResCAM \cite{HiResCAM}.
%In particular, we test the 3D saliency technique Saliency Tubes \cite{saliency_tubes}, and we extend three popular 2D saliency maps techniques for use with 3D data and models, including \textit{i}) Grad-CAM \cite{Grad-CAM}, \textit{ii}) Grad-CAM++ \cite{Grad-CAM++}, and \textit{iii}) HiResCAM \cite{HiResCAM}. 
Our findings confirm that, often, saliency methods explicitly tailored for 3D data outperform 3D extensions of 2D saliency methods. This underscores the importance of dedicated research in 3D-specific saliency methods. 
However, our investigation reveals a significant performance gap across all tested methods when compared to results obtained from 2D data, even for dimensionality-independent metrics such as the \texttt{F1} score. This discrepancy raises concerns about the adequacy of current saliency methods, both extensions of 2D methods and 3D-specific ones, in explaining the workings of 3D CNNs.
%However, our investigation unveils a notable discrepancy in the performance of all tested methods when compared to results achieved on 2D data, even for figures of merit such as the F1 score, that are independent of dimensionality. This raises concerns about the adequacy of current saliency methods, both 2D and 3D, to explain 3D CNNs.
%the performance of the extensions to Grad-CAM and Grad-CAM++ matches the performance of saliency methods explicitly designed for 3D data across all the evaluation metrics. Moreover, our experiments uncover a potentially concerning limitation of 3D saliency methods to explain 3D CNNs, as demonstrated by their weak performance. 

SE3D is the first framework enabling rigorous evaluation of  3D saliency methods. Moreover, SE3D lays the foundations for improving 3D CNN explanations, and for achieving more accurate WSOL and WSSS solutions, which in turn can drastically reduce the cost of gathering 3D localization and segmentation annotations. We publicly release our code and evaluation benchmark, including our implementation of 3D extensions of popular 2D saliency methods, metrics, model weights, and data processing code, at \url{https://github.com/Nexer8/SE3D}.

\section{Saliency Methods}
\label{sec:background}
%Our work proposes a framework for evaluating saliency methods for 3D CNN models, focusing on the medical domain. In this section, we present previous notable efforts related to saliency maps in medical imaging and their evaluation.
%In recent years, much emphasis has been placed on the use of artificial intelligence for medical image processing. For example, deep learning models have been used to diagnose Covid-19 based on CT scans \cite{CTCovidDiagnosis} and to segment tumors in MRI images \cite{MRIBrainSeg}. Most often here, the deep learning architecture of choice is the convolutional neural network (CNN) \cite{AlexNet}. While vision transformers (ViT) \cite{ViT} have recently surpassed the performance of CNNs in various tasks \cite{swin_transformer, detr, mask2former}, the ViT architecture does not allow the straightforward usage of saliency methods (Section \ref{sec:medical_saliency}). Since our aim is to evaluate saliency methods, we limit our study to CNN models. Furthermore, we examine saliency methods for a binary 3D CNN classifier, addressing tasks such as determining whether a 3D image contains a tumor or not.

%\subsection{Saliency Maps in Medical Imaging}
%\label{sec:medical_saliency}
The most common techniques employed to explain CNNs are based on saliency maps (Figure~\ref{fig:teaser}). 
A saliency map $s_\mathbf{x}$ for an input image $\mathbf{x}$ is a heatmap, with the same spatial dimensions as $\mathbf{x}$, that explains the decision of a model $\mathcal{M}$ with respect to a class $\lambda$. In practice, the values of $s_\mathbf{x}$ are larger in regions that most positively contributed to $\mathcal{M}$'s output for class $\lambda$.
% that highlights the regions of the image that positively contributed to the output of a model $\mathcal{M}$ with respect to a class $\lambda$ (Figure~\ref{fig:teaser}). 
% The values of the saliency map are larger in locations that most influenced the model's decision.
%Saliency maps are heatmaps of the same size of the image that highlight the regions of the image that most contributed to a specific model outcome with respect to a particular class $\lambda$ (Figure~\ref{fig:teaser}). 
% For a 2D image, we formalize saliency methods as a function:
We formalize saliency methods as:
%A saliency method $\mathcal{S}$ generates saliency map $s_\mathbf{x}$ to explain the decisions of a model $\mathcal{M}$
%can be formalized as a function:
% \vspace{-3pt}
% $$\mathcal{S}_{\mathcal{M}}^{\lambda}(\mathbf{x}): \mathbb{R}^{W,H,C} \rightarrow \mathbb{R}^{W,H}, \vspace{-4pt}$$ 
% and in the 3D case:
\vspace{-5pt}
$$\mathcal{S}_{\mathcal{M}}^{\lambda}(\mathbf{x}): \mathbb{R}^{W,H,D,C} \rightarrow \mathbb{R}^{W,H,D},\vspace{-5pt}$$
where depth $D=1$ in the 2D case and $C$ is the number of channels.

%, where $W,H$ are the spatial dimensions of the image and $C$ is the number of channels. 
CAM \cite{CAM} is arguably the most famous saliency method, and it leverages the activations at the last convolutional layer to compute a heatmap. This seminal work sparked a substantial research effort, giving rise to methods such as Grad-CAM \cite{Grad-CAM}, Grad-CAM++ \cite{Grad-CAM++}, HiResCAM \cite{HiResCAM}, and RISE \cite{RISE}, which have also been employed in medical imaging~\cite{gradcam_medical_imaging, cams_medical_imaging}.
Saliency maps have been applied both as explainability tools and for solving the Weakly Supervised Object Localization (WSOL) and Weakly Supervised Semantic Segmentation (WSSS) problems, with the latter being of particular interest in the medical field, where pixel-wise annotations are costly.

Despite their success, saliency methods have primarily been limited to explaining 2D CNNs. Attempts have been made at extending Grad-CAM \cite{Grad-CAM} to compute 3D explanations \cite{3d_gradcam1, 3d_gradcam2, chakraborty2020detection}, but due to the lack of a benchmark, these have never been rigorously evaluated. 
A few saliency methods have been designed for 3D CNNs \cite{gu2022xc, saliency_maps_for_segmentation_models, 3d_occlusion, gp_unet}, however, \cite{saliency_maps_for_segmentation_models, gp_unet} are restricted to specific CNN architectures, \cite{3d_occlusion} is only intended for videos, and \cite{gu2022xc} does not pass the sanity checks for saliency maps \cite{Saliency_Checks}.
To the best of our knowledge, Saliency Tubes \cite{saliency_tubes} and Respond-CAM \cite{Respond-CAM} are the only 3D saliency methods that are general enough to explain models in critical fields such as medical imaging, as they can be applied to 3D imaging data without any modification. Despite this, neither Saliency Tubes nor Respond-CAM have been rigorously evaluated. We propose the first rigorous framework for quantitatively assessing the performance of 3D saliency methods.

%, we extend a few popular 2D saliency methods to 3D data, and we compare them to 3D.

%, even though it has not been quantitatively evaluated. In our experiments, we propose the first rigorous framework for quantitatively assessing the performance of 3D saliency methods, extend some popular 2D saliency methods to 3D data, and compare these with Saliency Tubes.

\vspace{1mm}
\noindent\textbf{Evaluating 2D Saliency Methods:} %\label{sec:background_metrics}
In the 2D setting, saliency methods are evaluated by measuring their performance on the WSOL and WSSS tasks~\cite{WSOLRight}. The intuition behind this practice is that a good saliency map should focus on the object that is being explained, resulting in good localization and segmentation performance. 

%This quantitative evaluation is crucial as there is no direct way to assess the model's explainability capabilities. 
For 2D saliency methods, the most popular evaluation framework is proposed by 
Choe et al. \cite{WSOLRight}. In \cite{WSOLRight}, the authors propose the two metrics, \textbf{maximal box accuracy} (\texttt{MaxBoxAcc}) and \textbf{pixel average precision} (\texttt{PxAP}), to measure WSOL and WSSS performance, respectively.
%propose as evaluation metrics for WSOL and WSSS the \textbf{maximal box accuracy} (\texttt{MaxBoxAcc}) and \textbf{pixel average precision} (\texttt{PxAP}).
%which we detail in what follows, as these have inspired our metrics for 3D images. All the metrics 
These metrics are assessed over a set of annotated test images $\mathbf{X} = \{\mathbf{x}_1 \dots \mathbf{x}_N\}$, and their respective saliency maps $s_{\mathbf{x}}\,, \forall \mathbf{x} \in \mathbf{X}$. The first metric, \texttt{MaxBoxAcc}, measuring WSOL performance, is defined as: 
\begin{equation}
    \label{eq:maxboxacc}
    \texttt{MaxBoxAcc} = \max_{\tau} \frac{1}{N} \sum_{\mathbf{x} \in \mathbf{X}} \left[IoU(box(s_{\mathbf{x}}, \tau), B(\mathbf{x})) \geq \delta\right],
\end{equation}
where $box(s_{\mathbf{x}}, \tau)$ is the bounding box of the largest connected component retrieved from $s_{\mathbf{x}}$ after thresholding at $\tau$ ($\{s_{\mathbf{x}} \geq \tau \}$), and $B(\mathbf{x})$ is the ground truth bounding box in $\mathbf{x}$. Each term in the summation equals $1$ if and only if the $IoU$ between $box(s_{\mathbf{x}}, \tau)$ and $B(\mathbf{x})$ is above a chosen threshold $\delta$, typically set to $0.5$. 

Computing WSOL performance via \texttt{MaxBoxAcc} requires only bounding box annotations.
Instead, to assess WSSS performance, segmentation masks $m_{\mathbf{x}}$ are required, and \textit{pixel average precision} (\texttt{PxAP}) measures how well the saliency maps align with the masks. 
% To compute WSOL performance via \texttt{MaxBoxAcc}, bounding-box annotations are sufficient. 
% Instead, to compute WSSS performance, segmentation masks $m_{\mathbf{x}}$ are required.
% In this case,
% % \texttt{MaxBoxAcc} requires only bounding-box annotations.
% % When segmentation masks $m_{\mathbf{x}}$ are provided, instead, 
% \textit{pixel average precision} (\texttt{PxAP}) measures how well the saliency maps align with the masks. 
%it is possible to assess how well the saliency maps align with the segmentation ground truth by the pixel average precision. Given a saliency map $s_{\mathbf{x}_i}$, we denote its thresholded version as $\{s_{\mathbf{x}_i} \geq \tau \}$ and compare this against the ground truth segmentation pixels $\{m_{\mathbf{x}_i}=1 \}$. Then
\texttt{PxAP} is defined as the area under the pixel-wise precision-recall curve computed over the thresholded saliency map $\{s_{\mathbf{x}} \geq \tau \}$ at multiple thresholds $\tau = \tau_1 \dots \tau_l$:
\begin{flalign}
& \texttt{PxPrec}(\tau, x) = \frac{|\{s_{x} \geq \tau \} \cap \{m_{x}=1 \}|}{|\{s_{x} \geq \tau \}|}\,, && \\
&&& \mathmakebox[-63pt][r]{%
   \texttt{PxRec}(\tau, x) = \frac{|\{s_{x} \geq \tau \} \cap \{m_{x}=1 \}|}{|\{m_{x}=1 \}|}\,,
  \hspace{-1.8em}
} 
&& \\
&&& \mathmakebox[-150pt][r]{%
   \texttt{PxP}(\tau_l, \tau_{l-1}, x) = 
  \hspace{-1.8em}
}
&& \\
&&& \mathmakebox[-0pt][r]{%
   \texttt{PxPrec}(\tau_l, x)(\texttt{PxRec}(\tau_l, x) - \texttt{PxRec}(\tau_{l - 1}, x)),
  \hspace{-1.8em}
}
\notag
&& \\
&&& \mathmakebox[-72pt][r]{%
   \texttt{PxAP} = \frac{1}{N} \sum_l \sum_{\mathbf{x} \in \mathbf{X}}\texttt{PxP}(\tau_l, \tau_{l-1}, \mathbf{x}).
  \hspace{-1.8em}
}
&& \label{eq:pxap}
\end{flalign}

% \begin{equation}
%     \texttt{PxAP} = \frac{1}{N} \sum_l \sum_{\mathbf{x}_i \in \mathbf{X}} \texttt{PxPrec}(\tau_l, \mathbf{x}_i)(\texttt{PxRec}(\tau_l, \mathbf{x}_i) - \texttt{PxRec}(\tau_{l - 1}, \mathbf{x}_i)).
% \end{equation}
% but wsss requires segmentation masks
In practice, given a trained classification model $\mathcal{M}$ and an input image $\mathbf{x}$, the WSOL performance of the saliency method $\mathcal{S}_\mathcal{M}^\lambda(\mathbf{x})$ is assessed by measuring how much the bounding boxes of the thresholded heatmaps $\{s_{\mathbf{x}} \geq \tau \}$ align with the ground truth bounding box of the objects of class $\lambda$ in the images. Similarly, for WSSS, performance is measured by how much the thresholded heatmaps overlap with the ground truth segmentations for objects of class $\lambda$.

Evaluating \texttt{MaxBoxAcc} and \texttt{PxAP} is only possible for datasets possessing \textit{both} image-level classification information and pixel-level segmentation. The former is required to train the classifier for which to compute saliency maps, and the latter is required to evaluate the saliency maps on WSOL/WSSS.
%over bounding boxes and binary masks. 
For 2D CNNs, datasets such as CUB-2011~\cite{CUB2011} have been previously utilized for evaluation~\cite{WSOLRight}.
However, there are no benchmarks to quantitatively assess saliency methods for 3D CNNs. Previous works have been limited to qualitative assessment~\cite{saliency_tubes} or synthetic data analysis~\cite{Respond-CAM}.
This is largely due to the lack of 3D imaging datasets possessing image-level and voxel-level annotations. In our work, we propose modifications to ShapeNet~\cite{shapenet}, ScanNet~\cite{scannet}, and BraTS~\cite{BraTS1} datasets to overcome this limitation, and design new metrics, thus creating rigorous and comprehensive benchmarks for 3D CNN saliency methods.

\section{SE3D}
\label{sec:methods}
%Our aim is to develop a framework to evaluate saliency methods for 3D CNNs.
We propose SE3D, a framework to assess saliency methods for 3D CNNs.
Given a classifier $\mathcal{M}$ and a dataset $\mathcal{D} = \{(\mathbf{x}_i, \lambda_i)\}$ composed of $N$ volumes $\mathbf{x}_1 \dots \mathbf{x}_N \in \mathbb{R}^{W,H,D,C}$, each pertaining to a class $\lambda_1 \dots \lambda_N \in \Lambda$, we evaluate the performance of a saliency method $\mathcal{S}_{\mathcal{M}}^{\lambda_i}(\cdot)$. 
In the following, we detail the proposed 
3D datasets
(Section~\ref{sec:dataset}),
saliency method evaluation metrics
(Sections~\ref{sec:wsol_metrics},~\ref{sec:wsss_metrics}),
and models (Section~\ref{sec:details_model}) used in our benchmarks and experiments.

%Given a classifier model $\mathcal{M}$ and a classification dataset $\mathcal{D} = \{(\mathbf{x}_i, \lambda_i)\}$ composed of $N$ volumes $\mathbf{x}_1 \dots \mathbf{x}_N \in \mathbb{R}^{W,H,D,C}$, each pertaining to a class $\lambda_1 \dots \lambda_N \in \Lambda$, we evaluate the performance of a saliency map  $s_{\mathbf{x}_i} = \mathcal{S}_{\mathcal{M}}^{\lambda_i}(\mathbf{x}_i)$ for a particular volume $\mathbf{x}_i$. 
%To do so, we develop a framework to assess the WSOL and WSSS performance of 3D saliency methods, and provide a benchmark dataset and model.

%we want to measure the capability of the saliency map of explaining the model. To do so, we develop an evaluation framework composed of a benchmark model and dataset, and propose metrics to evaluate 3D saliency maps.

\subsection{Datasets}
\label{sec:dataset}

By far the greatest challenge to enable evaluation of 3D saliency methods is the definition of benchmark datasets. As discussed in Section~\ref{sec:background}, a suitable dataset must be labeled both at sample-level, in order to train the model, and at voxel-level, to evaluate the localization accuracy of the saliency maps.
%As stated in Section~\ref{sec:background}, both image-level classification (to train classifiers) and pixel-level segmentation labels are needed. 
However, no such dataset is readily available for 3D imaging data. This is because classification datasets do not provide expensive segmentation information, while segmentation datasets only focus on \textit{positive} samples (e.g. a tumor segmentation dataset will not contain volumes with no tumor occurrences).
Simple workarounds, such as combining two segmentation datasets and differentiating between them, are also unsuitable. Indeed, the approach may result in ``clever Hans" phenomena~\cite{cleverhans}, as the model may learn to classify based on spurious dataset biases.

To overcome this issue, we introduce modifications to three 3D imaging datasets: ShapeNet~\cite{shapenet}, ScanNet~\cite{scannet}, and BraTS~\cite{BraTS1}. These were chosen to cover a wide spectrum of scenarios, from synthetic data (ShapeNet), to real-world scans (ScanNet), and a critical setting in medical imaging (BraTS). 
We detail the procedure to construct five datasets: \hyphtexttt{shapenet-binary}, \hyphtexttt{shapenet-pairs}, \hyphtexttt{scannet-isolated}, \hyphtexttt{scannet-crop}, and \hyphtexttt{brats-halves}, all of which feature both sample and voxel level annotations. 
For all the proposed datasets, we release data processing tools that enable to construct the modified datasets starting from the publicly available ones. More information is available at the project's repository.
%These are further discussed in the supplementary material.

\vspace{1mm}
\noindent \textbf{ShapeNet:}
ShapeNet~\cite{shapenet} is a dataset composed of $51\,300$ synthetic CAD models each belonging to one of $51$ common object classes, such as \texttt{table}. To construct a benchmark on this dataset, we propose a straightforward procedure, where two classes $\lambda_1, \lambda_2$ are randomly chosen, and all models not pertaining to these two classes are discarded. The result is a dataset suitable for binary classification. 

\begin{figure}[t]
    \centering
    \begin{tabular}{ccc}
        \begin{subfigure}{0.13\textwidth}
            \centering
            \includegraphics[width=\linewidth]{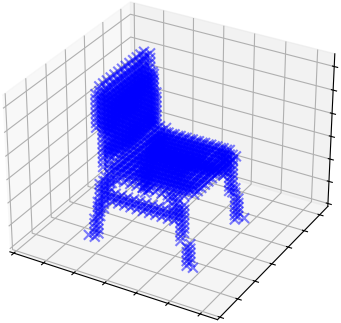}
            \caption{shapenet-binary}
            \label{fig:sample_shapenetbinary}
        \end{subfigure}
        &
        \begin{subfigure}{0.13\textwidth}
            \centering
            \includegraphics[width=\linewidth]{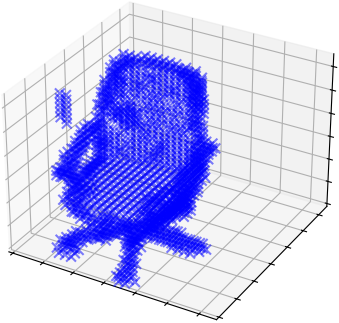}
            \caption{scannet-isolated}
            \label{fig:sample_scannetisolated}
        \end{subfigure}
        &
        \begin{subfigure}{0.13\textwidth}
            \centering
            \includegraphics[width=\linewidth]{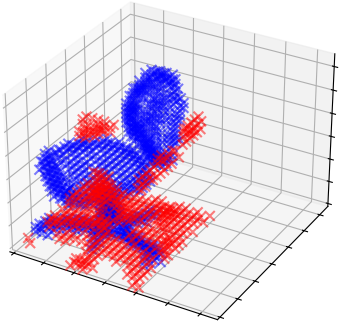}
            \caption{scannet-crop}
            \label{fig:sample_scannetcrop}
        \end{subfigure}
    \end{tabular}
    \caption{Samples for the proposed ShapeNet~\cite{shapenet} and ScanNet~\cite{scannet} variants. We display the {\color{blue} sample/segmentation mask} and the {\color{red} environment} voxels.
    \vspace{-5pt}
    }
    \label{fig:enter-label}
\end{figure}

For this first benchmark, we simplify the classification task as much as possible to achieve the best performance. Indeed, a model that has not learned to discriminate well between classes may return poor saliency maps, resulting in incorrect assessment of the saliency method. Furthermore, by training on voxelized versions of clean 3D models, we avoid ``clever Hans" phenomena~\cite{cleverhans}, ensuring that the only features the model can learn from pertain to the object to be classified.
In practice, we construct a dataset which we name ``\texttt{shapenet-binary}" composed of 3D volumes that display a single object. In this setting, localization metrics can assess whether the saliency maps accurately highlight the object or erroneously emphasize empty space. 
To do so, for each CAD model in shapenet, we extract binary 3D volumes with shape $32\times32\times32$, where each voxel is $1$ if it intersects the 3D model, and $0$ otherwise. The ground truth label is inherited from the CAD model, and the segmentation mask corresponds to the volume itself (Figure~\ref{fig:sample_shapenetbinary}).
% The resulting dataset, named ``\texttt{shapenet-binary}" (Figure~\ref{fig:sample_shapenetbinary}), contains only isolated objects. In this setting, localization metrics can assess whether the saliency maps accurately highlight the object or erroneously emphasize empty space. 

\begin{figure}
    \centering
    \begin{adjustbox}{max width=0.40\textwidth}
        \begin{tabular}{cc}
             \multirow{2}{*}{\makecell[c]{\includegraphics[width=0.2\linewidth]{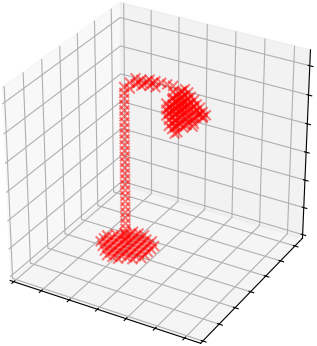} \\ $\mathbf{x}_N$}} & \multirow{2}{*}{\makecell[c]{\includegraphics[width=0.67\linewidth]{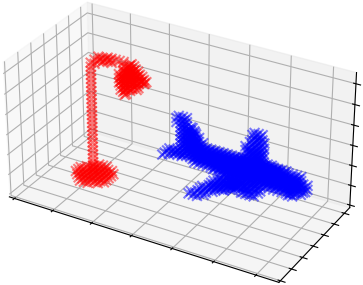} \\ $\mathbf{x}, o(\mathbf{x}) = 1$}} \\[65pt]
             \makecell[c]{\includegraphics[width=0.2\linewidth]{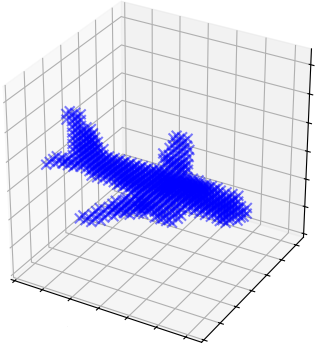} \\ $\mathbf{x}_C$} & 
        \end{tabular}
    \end{adjustbox}
    \caption{Generation of \texttt{shapenet-pairs} sample. The sample $\mathbf{x}$ is obtained by juxtaposing ShapeNet sample {\color{blue} $\mathbf{x}_C$} belonging to either $\lambda_1, \lambda_2$ and {\color{red} $\mathbf{x}_N$} belonging to $\Lambda \setminus \lambda_1, \lambda_2$.}
    \label{fig:shapenet-pairs}
\end{figure}

Saliency maps may also activate for irrelevant objects, rendering them non-class-discriminative~\cite{PV}.
%Previous studies~\cite{PV} have demonstrated that saliency maps can also activate for irrelevant objects, making them non-class-discriminative.
To verify whether this is the case for the tested methods, we introduce an additional ``\texttt{shapenet-pairs}" dataset comprising samples featuring two objects. Each sample in \texttt{shapenet-pairs} is designed to include an object $C$ relevant for classification and an irrelevant one $N$ (Figure~\ref{fig:shapenet-pairs}, left). 
%The objective is to discern whether the saliency map correctly identifies the relevant object.
In practice, we construct volumes with shape $64\times32\times32$, where each volume $\mathbf{x}$ is composed by adjoining two $32\times32\times32$ voxelized samples from ShapeNet, obtained in a similar fashion to \texttt{shapenet-binary}. Each pair is composed of one sample $\mathbf{x}_C$ belonging to one of two classes $\lambda_1, \lambda_2$ chosen at random, and one sample $\mathbf{x}_N$ from $\Lambda \setminus \{\lambda_1, \lambda_2\}$, where $\Lambda$ is ShapeNet's class set. 
We then concatenate $\mathbf{x}_C$ and $\mathbf{x}_N$ along their first dimension in a random order and obtain $\mathbf{x}$ (Figure~\ref{fig:shapenet-pairs}, right), which is labelled according only to $\mathbf{x}_C$. In this way, the object $\mathbf{x}_N$ acts as noise and should be ignored by a binary classifier over $\lambda_1, \lambda_2$.
Additionally, for evaluation purposes, we indicate for each $\mathbf{x}$ the position of $\mathbf{x}_C$ in the concatenation order via a function $o(\mathbf{x})$, which is $0$ if $\mathbf{x}_C$ is first and $1$ if it is second (Figure~\ref{fig:shapenet-pairs}, right).

\begin{figure}[t!]
    \centering
    \renewcommand\tabcolsep{1pt}
    \resizebox{0.54\textwidth}{!}{
        \begin{tabularx}{\textwidth}{|p{8pt}cccp{13pt}|p{8pt}ccccp{8pt}|}
            \cline{1-11}
            ~ & \makecell[c]{BraTS volume} & Left & Right & ~ & ~ & BraTS volume & Left & Right & ~ & \Tstrut 
            \\
            ~ & \multirow{2}{*}{\setlength{\fboxsep}{0pt}\setlength{\fboxrule}{0.2pt}\fbox{\includegraphics[width=0.257\textwidth,valign=T]{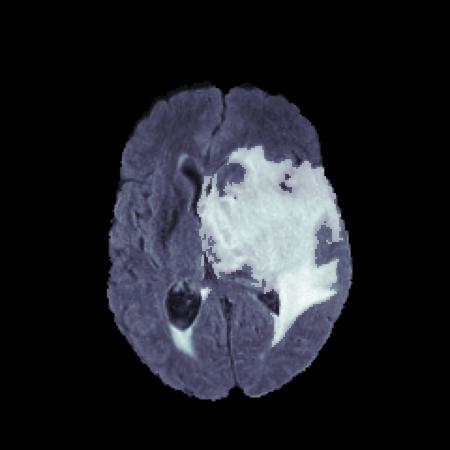}}} & 
            \setlength{\fboxsep}{0pt}\setlength{\fboxrule}{0.2pt}\fbox{\includegraphics[width=0.06\textwidth,valign=T]{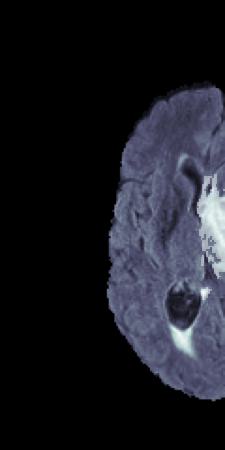}} &
            \setlength{\fboxsep}{0pt}\setlength{\fboxrule}{0.2pt}\fbox{\includegraphics[width=0.06\textwidth,valign=T]{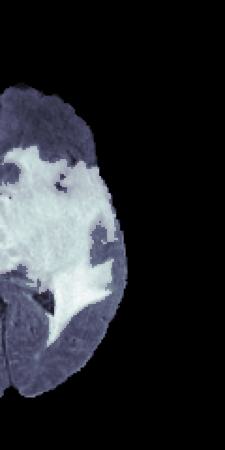}} & ~ & ~ &
            \multirow{2}{*}{\setlength{\fboxsep}{0pt}\setlength{\fboxrule}{0.2pt}\fbox{\includegraphics[width=0.257\textwidth,valign=T]{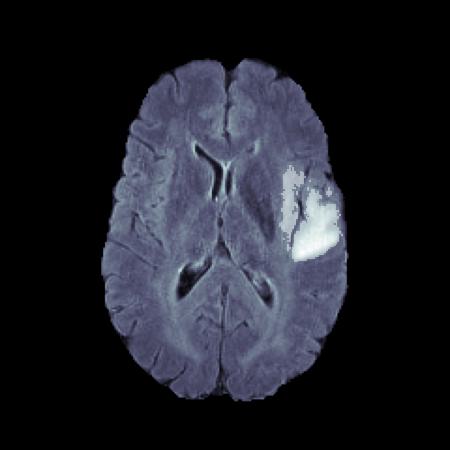}}} &
            \setlength{\fboxsep}{0pt}\setlength{\fboxrule}{0.2pt}\fbox{\includegraphics[width=0.06\textwidth,valign=T]{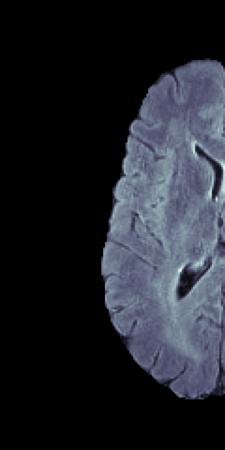}} &
            \setlength{\fboxsep}{0pt}\setlength{\fboxrule}{0.2pt}\fbox{\includegraphics[width=0.06\textwidth,valign=T]{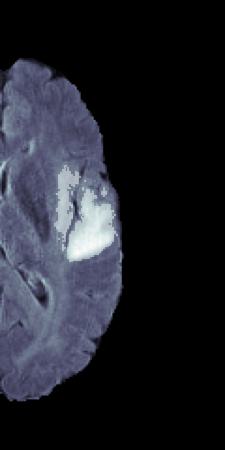}} &
            \put(-224,-30){\rotatebox[origin=c]{90}{Normal}} 
            \put(0,-30){\rotatebox[origin=c]{90}{Normal}} & ~ \\
            ~ & ~ & 
            \setlength{\fboxsep}{0pt}\setlength{\fboxrule}{0.2pt}\fbox{\includegraphics[width=0.06\textwidth,valign=T]{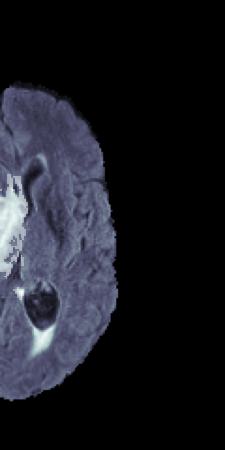}} &
            \setlength{\fboxsep}{0pt}\setlength{\fboxrule}{0.2pt}\fbox{\includegraphics[width=0.06\textwidth,valign=T]{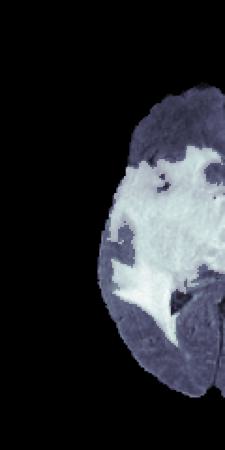}} &
            ~ & ~ & ~ &
            \setlength{\fboxsep}{0pt}\setlength{\fboxrule}{0.2pt}\fbox{\includegraphics[width=0.06\textwidth,valign=T]{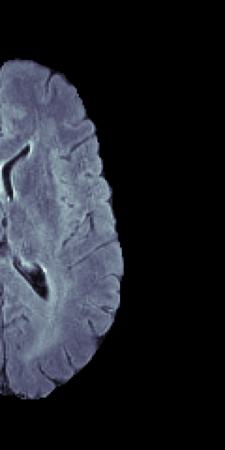}} &
            \setlength{\fboxsep}{0pt}\setlength{\fboxrule}{0.2pt}\fbox{\includegraphics[width=0.06\textwidth,valign=T]{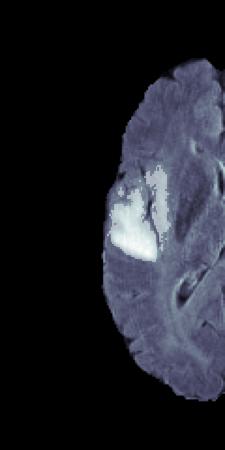}} &
            \put(-224,-30){\rotatebox[origin=c]{90}{Flipped}} 
            \put(0,-30){\rotatebox[origin=c]{90}{Flipped}} & ~ \\
    
            ~ & ~ & \small{\texttt{T}} & \small{\texttt{T}} & ~ & ~ & ~ & \small{\texttt{NT}} & \small{\texttt{T}} & ~ & ~ \Bstrut\\
            \cline{1-11}
        \end{tabularx}
        \put(-504,0){\rotatebox[origin=c]{90}{Single slices, top view}}
    }

    \caption{Generation of \texttt{brats-halves}. All BraTS samples contain tumors (highlighted in white). The hemispheres, however, could be devoid of tumor after splitting, and are labelled \texttt{tumor}~(\texttt{T}) or \texttt{no tumor}~(\texttt{NT}) accordingly.
    %are split and labeled as \texttt{tumor}~(\texttt{T}) or \texttt{no tumor}~(\texttt{NT}) based on the presence of the tumor.
    %We split each volume at the \textit{corpus callosum}, splitting the two brain hemispheres. If the tumor is present in both hemispheres (left pane), then both left and right samples will be classified as \texttt{tumor}~(\texttt{T}). If one hemisphere is healthy (right pane), it will be classified as \texttt{no tumor}~(\texttt{NT}). To avoid biases due to asymmetries in the hemispheres, we also flip each sample.
    \vspace{-10pt}
    }
    \label{fig:dataset_preparation}
\end{figure}

\vspace{1mm}
\noindent \textbf{ScanNet:}
ScanNet~\cite{scannet} is a dataset of densely annotated 3D reconstructions of more than $700$ indoor scenes, provided both as RGB-D sensor sequences and as meshes/point clouds.
We choose to benchmark on ScanNet to assess the performance of saliency methods in a more realistic scenario than ShapeNet, which only contains synthetic data.
Similarly to ShapeNet, we transform the datasets to consider only objects of two randomly chosen classes $\lambda_1, \lambda_2$. To construct a binary classification dataset, starting from ScanNet, we extract one volume for each object of class $\lambda_1, \lambda_2$ in the dataset in two ways (Figures~\ref{fig:sample_scannetisolated},~\ref{fig:sample_scannetcrop}), thus creating two datasets.
\begin{figure*}[t!]
    \renewcommand\tabcolsep{4pt}
    \resizebox{0.70\textwidth}{!}{
        \centering
        \hspace{0.05\textwidth}
        \begin{tabularx}{\textwidth}{Y|YY|Y|p{0pt}}
            \multicolumn{1}{c}{\large Measured against:} & \multicolumn{2}{c}{\large GT bounding box} & \multicolumn{1}{c}{\large GT segmentation} & \\ 
            \cline{2-4} ~ & \rule{0pt}{12pt}  \texttt{\large Max3DBoxAcc} & \texttt{\large Max3DBoxAccV2} & \texttt{\large VxAP} & \\
            \multirow[c]{3}{*}[30pt]{
                \setlength{\fboxsep}{0pt}\setlength{\fboxrule}{0.0pt}\fbox{\includegraphics[width=0.22\textwidth,align=T]{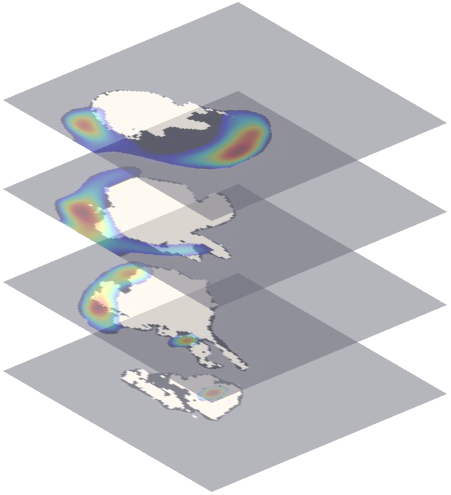}
                \put(-117,-10){\large Saliency + Segmentation}
                }
            }
            
             & \setlength{\fboxsep}{0pt}\setlength{\fboxrule}{0.0pt}\fbox{\includegraphics[width=0.22\textwidth,align=T]{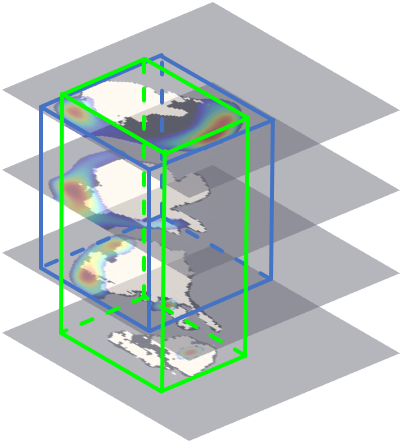}}
             \put(-20,5){\fontsize{12}{15}\selectfont {\textcircled{\raisebox{0.1pt} {$a$}}}}
             & \setlength{\fboxsep}{0pt}\setlength{\fboxrule}{0.0pt}\fbox{\includegraphics[width=0.22\textwidth,align=T]{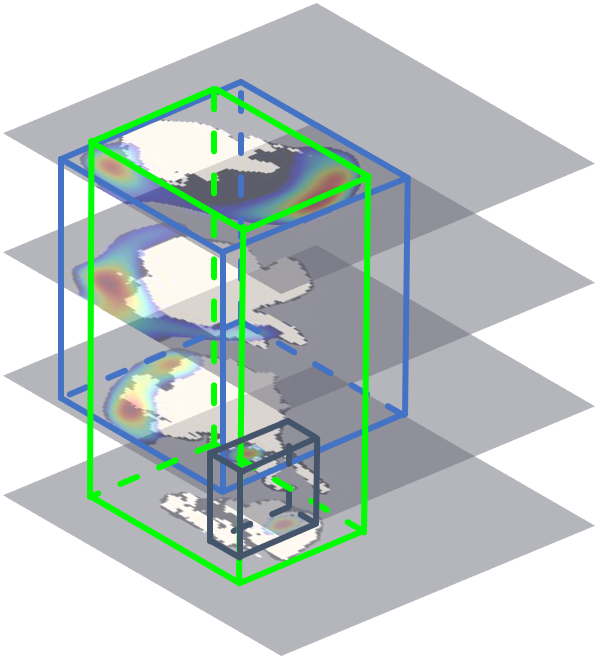}}
             \put(-20,5){\fontsize{12}{15}\selectfont {\textcircled{\raisebox{0.1pt} {$c$}}}}
             & \setlength{\fboxsep}{0pt}\setlength{\fboxrule}{0.0pt}\fbox{\includegraphics[width=0.22\textwidth,align=T]{images/metrics/vxap_withgt.png}
             \put(-20,5){\fontsize{12}{15}\selectfont {\textcircled{\raisebox{0.1pt} {$e$}}}}
             \put(-147,44){\color{red} *}} & \\
             \cline{2-4} ~ & \rule{0pt}{12pt} \texttt{\large MaxBoxAcc} & \texttt{\large MaxBoxAccV2} & \texttt{\large PxAP} & \\
             
             ~ &
             \setlength{\fboxsep}{0pt}\setlength{\fboxrule}{0.0pt}\fbox{\includegraphics[width=0.22\textwidth,align=T]{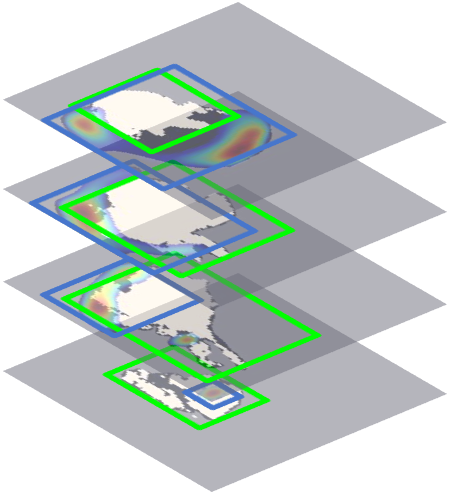}}
             \put(-20,5){\fontsize{12}{15}\selectfont {\textcircled{\raisebox{-0.4pt} {$b$}}}}
             & \setlength{\fboxsep}{0pt}\setlength{\fboxrule}{0.0pt}\fbox{\includegraphics[width=0.22\textwidth,align=T]{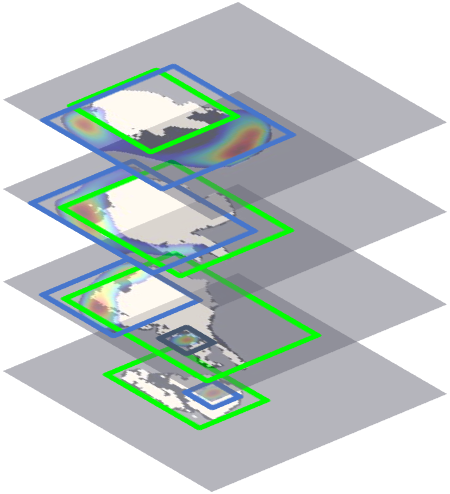}}
             \put(-20,5){\fontsize{12}{15}\selectfont {\textcircled{\raisebox{-0.4pt} {\fontsize{10}{11}\selectfont $d$}}}}
             & \setlength{\fboxsep}{0pt}\setlength{\fboxrule}{0.0pt}\fbox{\includegraphics[width=0.22\textwidth,align=T]{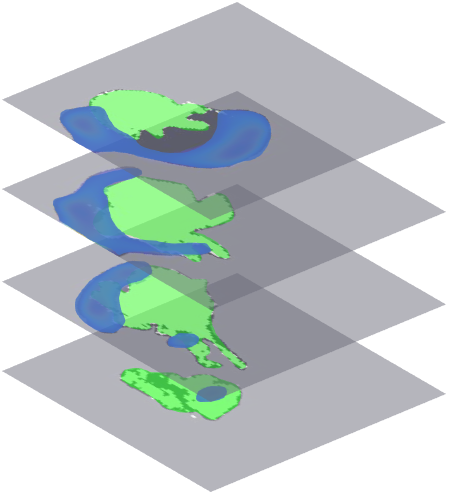}
             \put(-147,42){\large \color{red} *}} 
             \put(-20,5){\fontsize{12}{15}\selectfont {\textcircled{\raisebox{0.1pt} {\fontsize{8}{11}\selectfont $f$}}}}
             \\
             \cline{2-4}
        \end{tabularx}
        \put(10, 120){\Large {\color{green}{$\Box$}}\,\, Ground Truth (GT)}
        \put(10, 100){\Large {\color{blue}{$\Box$}}\,\, Predicted from saliency map}
        \put(12, 80){\Large {\color{red}{$*$}}\,\, 
        }
        \put(30, 27){\parbox{185pt}{\Large Highlights differences between 2D and 3D metrics. In the marked slices, a single connected volumetric component would be treated as two distinct components when analyzing the volume in a slice-wise manner.}}
        \put(-16,55){\colorbox{white}{\rotatebox{90}{\parbox{1cm}{\large \centering 3D}}}}
        \put(-16,-95){\colorbox{white}{\rotatebox{90}{\parbox{1cm}{\large \centering 2D}}}}
    }

    \caption{Extensions of the three metrics proposed by~\cite{WSOLRight} to 3D data. We visualize the computation of 2D and 3D metrics using an example 3D CT scan from BraTS taken as a whole volume (top) or as a series of slices (bottom). Both \texttt{Max3DBoxAcc} and \texttt{Max3DBoxAccV2} compare IoU between the bounding boxes of the ground truth and of the saliency map. However, \texttt{Max3DBoxAcc} only considers the largest connected component, while \texttt{Max3DBoxAccV2} matches each bounding box from the prediction to a box in the ground truth. In \textcircled{\raisebox{0.1pt} {$c$}}, \textcircled{\raisebox{-0.3pt} {$d$}}, only the \textcolor{lightblue}{blue box} matches the \textcolor{img_green}{green} ground-truth box, and not the \textcolor{bluegray}{blue-gray box}. \texttt{VxAP} compares the \textcolor{img_green}{GT} and the \textcolor{lightblue}{saliency map} in a voxel-wise fashion. %In the bottom row, we compare the proposed 3D metrics to 2D metrics applied to each slice of the volume.
    \vspace{-10pt}
    }
    \label{fig:metrics}
\end{figure*}
% The resulting dataset is suitable for binary classification. By choosing a similar setup to that used for ShapeNet, when choosing the same $\lambda_1, \lambda_2$ for both datasets, we enable interchangeability of the datasets between training and testing. In particular, this enables to train a model on the synthetic ShapeNet, thus avoiding ``clever Hans" phenomena, but testing it on the real-world ScanNet. Indeed, as we show in Section~\ref{sec:experiments}, the loss in performance when testing a ShapeNet-trained model on ScanNet is marginal.
%We extract volumes from ScanNet in two ways (Figures~\ref{fig:sample_scannetisolated},~\ref{fig:sample_scannetcrop}). 
% We propose two different ways to extract volumes from ScanNet's annotated point clouds (Figures~\ref{fig:sample_scannetisolated},~\ref{fig:sample_scannetcrop}). 
The first dataset, which we name ``\texttt{scannet-isolated}", is obtained by filtering each point cloud and only extracting the points that belong to the object, according to ScanNet labels. The second dataset, named ``\texttt{scannet-crop}", is obtained by considering all points falling within an axis-aligned bounding box that tightly contains the object, possibly including points from other objects in the environment (red points in Figure~\ref{fig:sample_scannetcrop}).

For both datasets, the point cloud subsets are voxelized to volumes with shape $32\times32\times32$, where each voxel is $1$ if at least one point in the point cloud intersects the grid region pertaining to the voxel, and $0$ otherwise.
For both \texttt{scannet-isolated} and \texttt{scannet-crop} samples, the segmentation mask corresponds to the voxelized point clouds filtered on the object instance, excluding environment points (Figures~\ref{fig:sample_scannetisolated},~\ref{fig:sample_scannetcrop}).
% for each sample are taken as the voxelized point clouds filtered as per \texttt{scannet-isolated}, thus only containing the object of interest (Figures~\ref{fig:sample_scannetisolated},~\ref{fig:sample_scannetcrop}).

\vspace{1mm}
\noindent \textbf{BraTS:}
To evaluate saliency methods in a critical setting such as medical imaging, we propose a third benchmark on the BraTS tumor segmentation dataset, composed of $\approx300$ CT scans with shape $240\times120\times155$ and $4$ channels. For each sample, a segmentation is provided for the tumor region.

Many medical imaging datasets exist, however, none provide both segmentation and classification information. While BraTS is no exception to this, the symmetry of the imaged brains enables us to propose a partitioning procedure and obtain a suitable dataset, which we name ``\texttt{brats-halves}".
The intuition is to split the imaged brains along the \textit{corpus callosum} joining the two brain hemispheres, as shown in Figure \ref{fig:dataset_preparation}.
After this partitioning, positive samples are comprised by those partitions that, according to the segmentation mask, do contain a tumor (marked as \texttt{T} in Figure \ref{fig:dataset_preparation}). Negative samples, instead, are comprised by those partitions that are devoid of tumors (marked as \texttt{NT} in Figure \ref{fig:dataset_preparation}).
Negative and positive partitions can be used 
to train a binary classifier, which can be evaluated on WSOL/WSSS given the available tumor segmentation masks.
%for binary classification, and WSOL/WSSS evaluation can be performed on the available tumor segmentation masks.
%We identify two suitable tumor segmentation datasets in the literature: BraTS \cite{BraTS1} for brain and KiTS \cite{KiTs} for kidneys. For BraTS, we split volumes along the \textit{corpus callosum} joining the two brain hemispheres, as shown in Figure \ref{fig:dataset_preparation}. For KiTS, we split the volume in correspondence to the patient's sagittal plane, obtaining volume partitions each displaying the left or right kidney.
%We identify the BraTS brain tumor segmentation dataset as a suitable candidate, and we split the dataset volumes along the \textit{corpus callosum} joining the two brain hemispheres, as shown in Figure \ref{fig:dataset_preparation}.
%We identify three suitable tumor segmentation datasets in the literature: BraTS for brain, KiTS for kidneys, and medical decathlon's lung for lungs. \mw{Leave just BraTS for now.} For BraTS, we split volumes along the \textit{corpus callosum} joining the two brain hemispheres, as shown in Figure \ref{fig:dataset_preparation}. For KiTS and medical decathlon's lung, instead, we split the volume in correspondence to the patient's sagittal plane, obtaining volume partitions each displaying the left or right kidney/lung. 
Since the tumor could be located anywhere in the volume, splitting may result in exceptionally hard hemispheres containing a very small fraction of tumor voxels. Thus, we choose to remove from the dataset all samples that, after splitting, 
contain a fraction of tumor voxels $t$ in $0<t<0.003$, as measured from the GT segmentation.
Lastly, to further avoid biases due to small asymmetries between left and right hemispheres, we augment the dataset by randomly flipping volumes along the symmetry plane during training (Figure~\ref{fig:dataset_preparation}, Left/Right).
%We publicly release data processing code.
%For licensing reasons, we release scripts to prepare the modified dataset and redirect to the official website to download the BraTS dataset.

We have also attempted to include KiTS~\cite{KiTS21} and medical imaging decathlon ``lung"~\cite{medical_decathlon} datasets in our framework.
For these, we split volumes according to the patient's sagittal plane, obtaining partitions each displaying one of the two kidneys/lungs. Despite our efforts, we were not able to train adequate classification models to enable fair evaluation of saliency methods.
We hypothesize that the low performance is due to the task's complexity (due to small dataset size and small fraction of tumor voxels), as demonstrated by the low performance of state-of-the-art segmentation models on these datasets. In particular, the best available models achieve Dice scores of $0.68$ \cite{kits68} and $0.69$ \cite{lungs69} for KiTS and medical imaging decathlon ``lung", respectively. The model from \cite{kits85} reportedly achieves a Dice score of $0.85$ on KiTS, but we were not able to reproduce similar results, even when employing the authors' proposed libraries. Thus, concerning medical imaging, SE3D only considers the BraTS dataset.

\subsection{WSOL Metrics}
%\subsection{Performance Metrics}
\label{sec:wsol_metrics}
%\label{sec:methods_benchmark}
We propose to evaluate saliency methods on two tasks. The first is Weakly Supervised Object Localization (WSOL), which measures localization accuracy at the bounding-box level. 
While it is possible to treat a 3D image as a series of 2D slices, thus using the same metrics as in~\cite{WSOLRight} such as \texttt{MaxBoxAcc} (\ref{eq:maxboxacc}), this approach would fail to capture the 3D shape of the objects, and may result in connected 3D components being separated when split in 2D slices, as shown in the different boxes in the top and bottom row of Figure~\ref{fig:metrics}.
Thus, we propose a 3D extension of \texttt{MaxBoxAcc}, enabling comparisons with previous results on 2D benchmarks.

%As stated in Section \ref{sec:background}, Choe et al. \cite{WSOLRight} propose to evaluate saliency maps on the WSOL and WSSS tasks by measuring \texttt{MaxBoxAcc} and \texttt{PxAP}. These two metrics, however, are designed for 2D saliency maps. While it is possible to treat a 3D image as a series of 2D slices, this approach would fail to capture the 3D shape of the objects to be localized or segmented (Figure~\ref{fig:metrics}). Instead,
%As described in Section \ref{sec:medical_saliency}, Choe et al. \cite{WSOLRight} proposes to evaluate saliency maps on the WSOL and WSSS taks by measuring \texttt{MaxBoxAcc} and \texttt{PxAP}. These two metrics, however, are designed for 2D saliency maps. While it is possible to treat a 3D image as a series of 2D slices, this approach would fail to capture additional information entailed by the 3D nature of the input and saliency map. 
%To evaluate 3D saliency methods,
%\noindent\textbf{Metrics: }
%we propose extensions to the \texttt{MaxBoxAcc} and \texttt{PxAP} metrics presented in  Choe et al. \cite{WSOLRight} to assess 3D saliency methods.
In the following, $box(s_{\mathbf{x}}, \tau)$ refers to the 3D bounding box surrounding the largest connected component of the thresholded saliency map $\{s_\mathbf{x} \geq \tau\}$. Further, $B(\mathbf{x})$ refers to the 3D bounding box around the largest component of the ground truth segmentation mask $m_{\mathbf{x}}$, and $\delta$ refers to the $IoU$ threshold indicating the desired level of accuracy. The metrics are computed over the dataset volumes $\mathbf{X} = \{\mathbf{x}_1,\dots,\mathbf{x}_N\}$.
%We thus extend the three metrics (Figure \ref{fig:metrics}):

\vspace{1mm} \noindent \textbf{Maximal 3D Box Accuracy} (\texttt{Max3DBoxAcc}) is the direct 3D extension of \texttt{MaxBoxAcc} (\ref{eq:maxboxacc}), and measures the maximal $IoU$ between the 3D bounding box around the largest connected component of the saliency map and the largest 3D bounding box of the ground truth (Figure \ref{fig:metrics}, \textcircled{\raisebox{0.1pt} {$a$}}). For a saliency map thresholded at $\tau$ and $IoU$ threshold $\delta$, we define \texttt{3DBoxAcc}~as:
\begin{equation}
    \texttt{3DBoxAcc}(\tau, \delta) = \frac{1}{N} \sum_{\mathbf{x} \in \mathbf{X}} \left[IoU(box(s_{\mathbf{x}}, \tau), B(\mathbf{x})) \geq \delta\right].
\end{equation}
\texttt{Max3DBoxAcc} is computed by maximizing \texttt{3DBoxAcc} over $\tau$, as $\texttt{Max3DBoxAcc}(\delta) = \max_{\tau} \left(\texttt{3DBoxAcc}(\tau, \delta)\right)$. This metric differs from the slice-wise 2D \texttt{MaxBoxAcc} (\ref{eq:maxboxacc}), since \texttt{Max3DBoxAcc} considers a single bounding box for the entire volume, rather than different ones for each slice (Figure \ref{fig:metrics}, \textcircled{\raisebox{-0.9pt} {$b$}}).

\vspace{1mm} \noindent \textbf{Maximal 3D Box Accuracy V2} (\texttt{Max3DBoxAccV2}) 
is a variant of \texttt{Max3DBoxAcc} that considers all 3D connected components of the saliency map and all the 3D ground truth bounding boxes, when multiple ones are present (Figure \ref{fig:metrics}, \textcircled{\raisebox{0.1pt} {$c$}}, \textcircled{\raisebox{-0.9pt} {$d$}}). It is defined as:
%when $\mathbf{x}$ contains multiple target boxes, \texttt{Max3DBoxAccV2} considers all connected components of the saliency map and all the 3D ground truth bounding boxes. Similarly to Choe et al. \cite{WSOLRight}, we define:
\begin{flalign}
% \vspace{-10pt}
& \texttt{3DBoxAccV2}(\tau, \delta) = && \\
&&& \mathmakebox[-0pt][r]{%
   \frac{1}{N} \sum_{\mathbf{x} \in \mathbf{X}} \max_{j,\,k} \left[IoU(box^j(s_{\mathbf{x}}, \tau), B^k(\mathbf{x})) \geq \delta\right] ,
  \hspace{-1.8em}
  % \vspace{-4pt}
}
\notag
\end{flalign}
where $box^j(s_{\mathbf{x}}, \tau)$ is the 3D bounding box around the $j$-th 3D connected component of the saliency map, and $B^k(\mathbf{x})$ is the $k$-th 3D bounding box of the ground truth object(s).
Due to maximization over all indexes $j,k$, the metric returns $1$ when there is at least one pair $(box^j(s_{\mathbf{x}}, \tau), B^k(\mathbf{x}))$ with overlap greater than $\delta$. Similarly to \texttt{Max3DBoxAcc}, this metric is maximized over all $\tau$, however, differently than \texttt{Max3DBoxAcc}, \texttt{Max3DBoxAccV2} is computed by averaging over a set of thresholds $\delta \in \Delta$ to account for varying levels of localization accuracy:
\begin{equation}
    \text{\texttt{Max3DBoxAccV2}} = \max_{\tau}(\frac{1}{|\Delta|}\sum_{\delta \in \Delta}{\text{\texttt{3DBoxAccV2}}(\tau,\delta)}).
\end{equation}
This metric differs from averaging the 2D \texttt{MaxBoxAccV2} over all slices, since connected 3D components may constitute multiple differently shaped components in each 2D slice (Figure \ref{fig:metrics}, \textcircled{\raisebox{-0.9pt} {$d$}}).

\subsection{WSSS Metrics}
\label{sec:wsss_metrics}
The second task for evaluating saliency methods is Weakly Supervised Semantic Segmentation (WSSS), which measures localization accuracy at the voxel level. 
We extend \texttt{PxAP}~\cite{WSOLRight} (\ref{eq:pxap}) to enable comparison with previous benchmarks for 2D CNN saliency maps. We also propose novel metrics: \textbf{Maximal F1 Score} (\texttt{MaxF1}), \textbf{Precision at Optimal F1} (\texttt{Prec@F1$\tau$}), and \textbf{Recall at Optimal F1} (\texttt{Rec@F1$\tau$}), that provide an absolute measure of how much the saliency map covers the object of interest and how much it is dispersed throughout the volume.
Additionally, we propose the \textbf{Mass Concentration} (\texttt{MC}) metric to evaluate whether the saliency map focuses on irrelevant objects.
%texttt{MaxF1}, \texttt{Prec@F1\_$\tau$}, \texttt{Rec@F1\_$\tau$}, and \texttt{MC}.

\vspace{1mm} \noindent \textbf{Voxel Average Precision} (\texttt{VxAP}) is the 3D extension of \texttt{PxAP} (\ref{eq:pxap}) to the 3D case, and measures the average precision and recall of the saliency map with respect to the ground truth segmentation mask ${m_{x}}$. These are defined as:
\begin{flalign}
& 
    \label{eq:pxprec} \texttt{VxPrec}(\tau, x) = \frac{|\{s_{x} \ge \tau\} \cap \{m_{x} = 1\}|}{|\{s_{x} \ge \tau\}|} , && \\
&&& 
    \label{eq:pxrec} \mathmakebox[-62pt][r]{%
   \texttt{VxRec}(\tau, x) = \frac{|\{s_{x} \ge \tau\} \cap \{m_{x} = 1\}|}{|\{m_{x} = 1\}|}\,.
   \hspace{-1.8em}
}
&& \\
&&& \mathmakebox[-149pt][r]{%
   \texttt{VxP}(\tau_l, \tau_{l-1}, x) = 
  \hspace{-1.8em}
}
&& \\
&&& \mathmakebox[-0pt][r]{%
   \texttt{VxPrec}(\tau_l, x)(\texttt{VxRec}(\tau_l, x) - \texttt{VxRec}(\tau_{l - 1}, x)),
  \hspace{-1.8em}
}
\notag
&& \\
&&& \mathmakebox[-71pt][r]{%
   \texttt{VxAP} = \frac{1}{N} \sum_l \sum_{\mathbf{x} \in \mathbf{X}}\texttt{VxP}(\tau_l, \tau_{l-1}, \mathbf{x}).
  \hspace{-1.8em}
}
% \vspace{-3pt}
\end{flalign}
% \begin{equation}
% \hspace{-10pt}
%     \texttt{VxPrec}(\tau, x) = \frac{|\{s_{x} \ge \tau\} \cap \{m_{x} = 1\}|}{|\{s_{x} \ge \tau\}|} , \quad
%     \texttt{VxRec}(\tau, x) = \frac{|\{s_{x} \ge \tau\} \cap \{m_{x} = 1\}|}{|\{m_{x} = 1\}|}\,.
% \end{equation}
% The \textbf{voxel average precision}, \texttt{VxAP}, is then defined as the area under the precision-recall curve at various levels of $\tau = \tau_1, \dots \tau_L$:
% \begin{equation}
%     \texttt{VxAP} = \frac{1}{N} \sum_l \sum_{\mathbf{x}_i \in \mathbf{X}} \texttt{VxPrec}(\tau_l, \mathbf{x}_i)(\texttt{VxRec}(\tau_l, \mathbf{x}_i) - \texttt{VxRec}(\tau_{l - 1}, \mathbf{x}_i)).
% \end{equation}
%To ensure that the final metric takes into account all the segmentation classes, we average the \texttt{VxAP} over all the classes of the ground truth segmentation mask. We call this metric \textbf{mean voxel average precision}, \texttt{VxAP}. \lo{what??} \mw{Check Teams.}
\texttt{VxAP} is a generalization of \texttt{PxAP}. When all volume slices have identical dimensions, it is possible to compute \texttt{VxAP} as the average \texttt{PxAP} over all slices ($\texttt{VxAP}(s_\mathbf{x}) \equiv \frac{1}{D} \sum_{i\in\{1\dots D\}}{\texttt{PxAP}(s_\mathbf{x}[:,:,i])}$ where $D$ is the volume's depth dimension) (Figure \ref{fig:metrics}, \textcircled{\raisebox{0.1pt} {$e$}}, \textcircled{\fontsize{7}{9}\selectfont \raisebox{-0.1pt} {$f$}}).
%the average \texttt{PxAP} computed over all slices is equivalent to the \texttt{VxAP} computed over the whole volume .
% \texttt{VxAP} is equivalent to the average \texttt{PxAP} over all slices ($\texttt{VxAP}(s_\mathbf{x}) \equiv \frac{1}{D} \sum_{i\in\{1\dots D\}}{\texttt{PxAP}(s_\mathbf{x}[:,:,i])}$) where $D$ is the volume's depth dimension.
This is because the average slice-wise \texttt{PxAP} is the average of the averages of same-sized groups.
In our experiments, samples are cuboidal, thus all volume slices have identical dimensions and this holds true. We still define \texttt{VxAP} to enable compatibility with future methods, where the assumption of identically sized slices could potentially not hold.

%\vspace{1mm}
%The introduced metrics \texttt{Max3DBoxAcc}, \texttt{Max3DBoxAccV2}, and \texttt{VxAP} facilitate the comparison of results on 3D data with previous benchmarks, such as the widely adopted one proposed by Choe et al.~\cite{WSOLRight}. While these metrics serve as valid tools for evaluating localization performance, the dependence of \texttt{Max3DBoxAcc} and \texttt{Max3DBoxAccV2} on threshold $\delta$ and quantization to largest connected components, as well as the aggregation over all $\tau$ for \texttt{VxAP}, introduce complexities that hinder an intuitive understanding of the metrics' values.
%To address this limitation, we introduce three novel metrics, \textbf{Maximal F1 Score} (\texttt{MaxF1}), \textbf{Precision at Optimal F1} (\texttt{Prec@F1$\tau$}), and \textbf{Recall at Optimal F1} (\texttt{Rec@F1$\tau$}), that provide an absolute measure of how much the saliency map covers the object of interest and how much it is dispersed throughout the volume.
%is dispersed outside of the region of interest, and the extent to which the saliency map covers the object. 
%These metrics aim to offer a more intuitive and comprehensive assessment of saliency map characteristics in the context of 3D data analysis.

\vspace{1mm}
\noindent \textbf{Maximal F1 Score} (\texttt{MaxF1}) is the sample-wise average F1 score between the thresholded saliency maps $\{s_{\mathbf{x}} \geq \tau \}$ and the ground truth segmentation masks $m_{\mathbf{x}}$, computed at the optimal $\tau$:
\begin{equation}
    \text{\texttt{F1}}(\tau) = \frac{1}{N} \sum_{\mathbf{x} \in \mathbf{X}}\frac{2 \times \text{\texttt{VxPrec}}(\tau, \mathbf{x}) \times \text{\texttt{VxRec}}(\tau, \mathbf{x})}{\text{\texttt{VxPrec}}(\tau, \mathbf{x}) + \text{\texttt{VxRec}}(\tau, \mathbf{x})},
\end{equation}
\begin{equation}
    \text{\texttt{MaxF1}} = \max_{\tau}(\text{\texttt{F1}}(\tau)),
\end{equation}
where \texttt{VxPrec} and \texttt{VxRec} are computed as in  (\ref{eq:pxprec}) and (\ref{eq:pxrec}), respectively.
% \texttt{MaxF1} indicates how sharply the saliency map separates the object of interest from other regions in the 3D volume. By maximizing over $\tau$, non-zero saliency values outside the region of interest are ignored, provided that they are lower than values inside this region.
% The term \texttt{MaxF1} measures the effectiveness of the saliency map in distinguishing the object of interest from the surrounding regions. Due to maximization process over $\tau$, the metric is independent to activations activations outside the region of interest, as long as the activations inside this specific region surpass those outside, ensuring a robust evaluation of the saliency map.
\texttt{MaxF1} measures the saliency map's ability to differentiate the object of interest from its surroundings. The maximization over $\tau$ ensures that the metric is only sensitive to the disparity between activation values within and outside the region of interest, maintaining robustness to non-zero activations outside the region of interest.

\vspace{1mm}
\noindent \textbf{Precision and Recall at Optimal F1} (\texttt{Prec@F1$\tau$}, \texttt{Rec@F1$\tau$}) are voxel precision and recall (\texttt{VxPrec} and \texttt{VxRec}) computed over the maps $\{s_{\mathbf{x}} \geq \tau_{F1} \}$ thresholded at $\tau_{F1}$ that maximizes the F1 score.

% \vspace{2mm}
% \noindent \textbf{Mask Coverage} (\texttt{MaskCov}) \gi{what about prec e rec at F1 max tau} is the average recall computed between the binarized saliency maps $\{s_{\mathbf{x}_i} \geq 0 \}$ and the ground truth segmentation masks:
% \begin{equation}
%     \text{\texttt{MaskCov}} = \frac{1}{N} \sum_{\mathbf{x}_i \in \mathbf{X}}\text{\texttt{VxRec}}(0, \mathbf{x}_i).
% \end{equation}
% \texttt{MaskCov} indicates what is the average portion of the object of interest which is covered by saliency values greater than $0$.
\vspace{1mm}
By analyzing \texttt{MaxF1}, \texttt{Prec@F1$\tau$}, and \texttt{Rec@F1$\tau$}, one can better grasp the behavior of the saliency method. High \texttt{MaxF1} indicates that the saliency map covers most of the object of interest and is not scattered. \texttt{Prec@F1$\tau$} and \texttt{Rec@F1$\tau$} can also give more precise insights, highlighting whether the saliency map activates only for portions of the region of interest (when \texttt{Rec@F1$\tau$} is low) or in other regions of the volume as well (when \texttt{Prec@F1$\tau$} is low).
% By analyzing \texttt{MaxF1}, \texttt{MaskCov}, and the relation between the two, it is easy to understand the behavior of the saliency method. A high \texttt{MaskCov} indicates that the saliency map activates on large portions of the object of interest, however, if paired with a low \texttt{MaxF1}, this indicates that the saliency maps are not sharp, and may cover less relevant regions. On the other end, a low \texttt{MaskCov} indicates that the saliency map is $0$ for large portions of the object of interest. However, if \texttt{MaxF1} is also high, this denotes that the saliency map is not dispersed, but rather it is concentrated on a particular region of the object of interest.

\vspace{1mm}
\noindent\textbf{Mass Concentration} (\texttt{MC})
is specifically designed for the paired \texttt{shapenet-pairs} dataset (Section~\ref{sec:dataset}, Figure~\ref{fig:shapenet-pairs}), and evaluates whether the saliency map focuses only on relevant objects.
Given a sample $\mathbf{x}$ of adjoined samples $\mathbf{x}_C$ ($C$lass target), and $\mathbf{x}_N$ ($N$oise), the trained classifier should only focus on the volume's half containing object $\mathbf{x}_C$.
To assess how much the saliency map reflects this, we measure the proportion of the mass of the saliency map that is located on the volume's half pertaining to $\mathbf{x}_C$:

\begin{equation}
    \texttt{MC} = \frac{1}{N} \sum_{\mathbf{x} \in \mathbf{X}} \frac{sum(s_\mathbf{x}[o(\mathbf{x})\times32:(o(\mathbf{x})+1)\times32,:,:])}{sum(s_\mathbf{x})},
    % \vspace{-3pt}
\end{equation}
% \begin{flalign}
%     \vspace{-10pt}
%     & \texttt{MC}(x) = && \\
%     &&& \mathmakebox[-0pt][r]{%
%        \frac{sum(s_x[o(x)\times32:(o(x)+1)\times32,:,:])}{sum(s_x)},
%       \hspace{-1.8em}
%       \vspace{-4pt}
%     }
%     \notag
% \end{flalign}
where $o(x) \in \{0,1\}$ indicates the concatenation order (Section~\ref{sec:dataset}).

%In summary, \texttt{Max3DBoxAcc} and \texttt{Max3DBoxAccV2} are to be used when bounding-box segmentation information is available, with \texttt{Max3DBoxAccV2} managing multiple target bounding boxes. If voxel-level segmentation is available, then \texttt{VxAP} can also be used to obtain a finer measurement. 
%In summary, \texttt{Max3DBoxAcc} and \texttt{Max3DBoxAccV2} are to be used when bounding-box segmentation information is available, with \texttt{Max3DBoxAccV2} managing multiple target bounding boxes. If voxel-level segmentation is available, then \texttt{VxAP} can also be used to obtain a finer measurement. Lastly, $\delta$ is a parameter that specifies the desired minimum overlap of bounding boxes. 
% \noindent
% Following Choe et al. \cite{WSOLRight}, we choose $\delta = 0.5$ when measuring \texttt{Max3DBoxAcc(V2)} and $\delta \in \{0.3, 0.5, 0.7\}$ when measuring \texttt{Max3DBoxAccV2}.

%\lo{we must comment how the failure cases of 2D maps are solved by our method, maybe with a picture if it fits} \mw{We can do that using the segmentation masks I sent you.}

%The performance of a saliency method on a chosen task and dataset will thus be measured by the metrics presented above. We elaborate on the details of the benchmark in Section \ref{sec:methods_benchmark}.

\begin{table*}[ht]
        \caption{Results of saliency map evaluation for 2D and 3D metrics. Since in our case \texttt{VxAP}$=$\texttt{PxAP}, we only report the former.}
        \label{tab:main_results}
        \begin{adjustbox}{max width=0.95\textwidth}
        \hspace{0.025\textwidth}
        \renewcommand{\arraystretch}{1.0}
        \aboverulesep=0ex
        \belowrulesep=0ex
        \begin{tabular}{l|l|cccccccc}
    \multicolumn{1}{c|}{\multirow{2}{*}{\makecell[c]{Dataset \\ $<$classes$>$ \\ (test accuracy)}}} & \multicolumn{1}{c|}{\multirow{2}{*}{Saliency Method}} & \multicolumn{2}{c|}{3D WSOL Metrics} & \multicolumn{4}{c|}{3D WSSS Metrics} & \multicolumn{2}{c}{2D WSOL Metrics} \\ \cline{3-10} 
    \multicolumn{1}{c|}{} & \multicolumn{1}{c|}{} & \makecell[c]{Max3DBoxAcc \\ ($\delta = 0.5$)} & \makecell[c]{Max3DBoxAccV2 \\ ($\delta \in \{0.3, 0.5, 0.7\}$)} & VxAP & MaxF1 & Prec@F1$\tau$ & Rec@F1$\tau$ & \makecell[c]{MaxBoxAcc \\ ($\delta = 0.5$)} & \makecell[c]{MaxBoxAccV2 \\ ($\delta \in \{0.3, 0.5, 0.7\}$)} \\
    \hline 
    \multirow{6}{*}{\rotatebox[origin=c]{0}{\makecell[c]{shapenet-binary \\ chair/table \\ ($0.989$)}}} & & & & & & & & & \\[-5pt]
     & Grad-CAM~\cite{Grad-CAM} & $\mathbf{0.38}$ & $\mathbf{0.42}$ & $0.11$ & $0.18$ & $0.11$ & $0.59$ & $0.14$ & $0.18$ \\
     & Grad-CAM++~\cite{Grad-CAM++} & $0.18$ & $0.35$ & $0.10$ & $0.18$ & $0.09$ & $\mathbf{1.00}$ & $0.10$ & $0.14$ \\
     & HiResCAM~\cite{HiResCAM} & $0.24$ & $0.34$ & $0.10$ & $0.19$ & $0.11$ & $0.60$ & $0.10$ & $0.14$ \\
     & Respond-CAM~\cite{Respond-CAM} & $0.11$ & $0.22$ & $0.10$ & $0.18$ & $0.10$ & $\mathbf{1.00}$ & $0.05$ & $0.09$ \\
     & SaliencyTubes~\cite{saliency_tubes} & $0.29$ & $0.40$ & $\mathbf{0.23}$ & $\mathbf{0.30}$ & $\mathbf{0.25}$ & $0.38$ & $\mathbf{0.19}$ & $\mathbf{0.29}$ \\[-5pt]
      & & & & & & & & & \\
    \hline 
    \multirow{6}{*}{\rotatebox[origin=c]{0}{\makecell[c]{scannet-isolated \\ chair/table \\ ($0.919$)}}} & & & & & & & & & \\[-5pt] 
     & Grad-CAM~\cite{Grad-CAM} & $0.39$ & $0.41$ & $0.05$ & $0.11$ & $0.07$ & $0.28$ & $0.02$ & $0.07$ \\
     & Grad-CAM++~\cite{Grad-CAM++} & $0.00$ & $0.01$ & $0.04$ & $0.10$ & $0.06$ & $0.24$ & $0.01$ & $0.05$ \\
     & HiResCAM~\cite{HiResCAM} & $0.00$ & $0.10$ & $0.05$ & $0.12$ & $0.08$ & $0.26$ & $0.01$ & $0.07$ \\
     & Respond-CAM~\cite{Respond-CAM} & $0.01$ & $0.18$ & $0.05$ & $0.11$ & $0.08$ & $0.20$ & $0.02$ & $0.08$ \\
     & SaliencyTubes~\cite{saliency_tubes} & $\mathbf{0.63}$ & $\mathbf{0.61}$ & $\mathbf{0.09}$ & $\mathbf{0.18}$ & $\mathbf{0.12}$ & $\mathbf{0.35}$ & $\mathbf{0.06}$ & $\mathbf{0.10}$ \\[5pt]
     \hline
    \multirow{6}{*}{\rotatebox[origin=c]{0}{\makecell[c]{scannet-crop \\ chair/table \\ ($0.917$)}}} & & & & & & & & & \\[-5pt] 
     & Grad-CAM~\cite{Grad-CAM} & $0.19$ & $0.31$ & $0.04$ & $0.07$ & $0.04$ & $0.37$ & $0.01$ & $0.10$ \\
     & Grad-CAM++~\cite{Grad-CAM++} & $0.00$ & $0.03$ & $0.03$ & $0.07$ & $0.04$ & $\mathbf{0.38}$ & $0.01$ & $0.04$ \\
     & HiResCAM~\cite{HiResCAM} & $0.01$ & $0.05$ & $0.04$ & $0.08$ & $0.04$ & $0.28$ & $0.01$ & $0.06$ \\
     & Respond-CAM~\cite{Respond-CAM} & $0.21$ & $0.27$ & $0.03$ & $0.07$ & $0.04$ & $0.35$ & $0.01$ & $0.08$ \\
     & SaliencyTubes~\cite{saliency_tubes} & $\mathbf{0.52}$ & $\mathbf{0.50}$ & $\mathbf{0.06}$ & $\mathbf{0.12}$ & $\mathbf{0.07}$ & $0.29$ & $\mathbf{0.03}$ & $\mathbf{0.27}$ \\[5pt]
     \hline
    % \multirow{6}{*}{\rotatebox[origin=c]{90}{\makecell[c]{scannet-pairs \\ ($0.967$)}}} & & & & & & & & & \\[-5pt] 
    %  & Grad-CAM~\cite{Grad-CAM} & $0.01$ & $0.06$ & $0.03$ & $0.09$ & $0.08$ & $0.11$ & $0.00$ & $0.02$ \\
    %  & Grad-CAM++~\cite{Grad-CAM++} & $0.01$ & $0.03$ & $0.02$ & $0.07$ & $0.08$ & $0.07$ & $0.00$ & $0.01$ \\
    %  & HiResCAM~\cite{HiResCAM} & $0.01$ & $0.03$ & $0.03$ & $0.09$ & $0.07$ & $0.12$ & $0.00$ & $0.02$ \\
    %  & Respond-CAM~\cite{Respond-CAM} & $0.01$ & $0.02$ & $0.02$ & $0.05$ & $0.03$ & $0.10$ & $0.00$ & $0.01$ \\
    %  & SaliencyTubes~\cite{saliency_tubes} & $0.37$ & $0.40$ & $0.20$ & $0.31$ & $0.30$ & $0.31$ & $0.08$ & $0.12$ \\[5pt]
    %  \hline
    \multirow{6}{*}{\rotatebox[origin=c]{0}{\makecell[c]{brats-halves \\ tumor/no tumor \\ ($0.796$)}}} & & & & & & & & & \\[-5pt] 
     & Grad-CAM~\cite{Grad-CAM} & $0.00$ & $0.00$ & $0.06$ & $0.13$ & $0.11$ & $0.18$ & $0.00$ & $0.03$ \\
     & Grad-CAM++~\cite{Grad-CAM++} & $0.00$ & $0.00$ & $0.09$ & $0.19$ & $0.15$ & $0.27$ & $0.03$ & $0.03$ \\
     & HiResCAM~\cite{HiResCAM} & $0.00$ & $0.00$ & $0.09$ & $0.18$ & $\mathbf{0.19}$ & $0.27$ & $0.01$ & $0.02$ \\
     & Respond-CAM~\cite{Respond-CAM} & $0.00$ & $0.00$ & $0.03$ & $0.10$ & $0.07$ & $0.16$ & $0.01$ & $0.01$ \\
     & SaliencyTubes~\cite{saliency_tubes} & $\mathbf{0.19}$ & $\mathbf{0.21}$ & $\mathbf{0.12}$ & $\mathbf{0.21}$ & $0.14$ & $\mathbf{0.40}$ & $\mathbf{0.12}$ & $\mathbf{0.13}$ \\[5pt]
     \hline
    \end{tabular}
        \end{adjustbox}
    \vspace{-10pt}
\end{table*}

\subsection{Models}\label{sec:details_model}
% \begin{figure}[t!]
%     \centering
%     \renewcommand\tabcolsep{0pt}
%     \resizebox{0.8\textwidth}{!}{
%         \begin{tabularx}{\textwidth}{YY}
%             %\texttt{Tumor percentage} & \texttt{Tumor percentage} \\ 
%             \setlength{\fboxsep}{0pt}\setlength{\fboxrule}{0.0pt}\fbox{\includegraphics[width=0.5\textwidth,align=T]{images/classification_performance/correct_classification.png} 
%              \put(55,-8){Tumor Percentage}
%              \put(-129,-8){Tumor Percentage}
%              \put(-188,40){\rotatebox[origin=c]{90}{Samples}}
%              \put(-140,85){\small{Correctly classified samples}}
%              }
            
%              & \setlength{\fboxsep}{0pt}\setlength{\fboxrule}{0.0pt}\fbox{\includegraphics[width=0.5\textwidth,align=T]{images/classification_performance/incorrect_classification.png}
%              \put(-188,40){\rotatebox[origin=c]{90}{Samples}}
%              \put(-145,85){\small{Incorrectly classified samples}}
%              } 
%              \\
%         \end{tabularx}
%     }

%     \caption{Distribution of samples that were correctly and incorrectly identified, depending on the portion of a volume that a tumor occupies. The majority of model mispredictions occur for cases where the tumor only makes up a small fraction of the volume.}
%     \label{fig:failure_cases}
% \end{figure}

We train 3D CNN classifiers on the proposed \texttt{shapenet-binary}, \texttt{shapenet-pairs}, \texttt{scannet-isolated}, \texttt{scannet-crop}, and \texttt{brats-halves} datasets. The classifiers share the same architecture, inspired by \cite{3d_cnn_model_architecture}, composed of $3$ convolutional blocks followed by GAP and one dense prediction layer.
For each dataset, we randomly select $80\%$ of samples for training and $20\%$ for testing. 
Classification performance is reported in Tables~\ref{tab:main_results},~\ref{tab:cammass}. For \texttt{shapenet-binary} and \texttt{shapenet-pairs}, test accuracy is over $95\%$, and over $90\%$ for \texttt{scannet-isolated} and \texttt{scannet-crop}. For the harder \texttt{brats-halves}, accuracy is $\approx 80\%$. This strong performance confirms that the models correctly identify patterns in the data relating to different classes, and thus that high-quality saliency maps could be generated.
%Further architecture and training details are discussed in the supplementary material.
Architecture and training details are available at the project's repository.
\section{Evaluation of 3D Saliency Methods}
\label{sec:experiments}
% \vspace{-3pt}

We test saliency methods using our evaluation framework. These include extensions to 2D methods and 3D-specific methods. 
For each method, we compute \texttt{Max3DBoxAcc}, \texttt{Max3DBoxAccV2}, \texttt{VxAP}, their 2D counterparts, and our proposed metrics \texttt{MaxF1}, \texttt{Prec@F1$\tau$}, \texttt{Rec@F1$\tau$}. On our hardware setup (AMD EPYC 7443P), even the most demanding metric \texttt{Max3DBoxAccV2} takes less than $1$s to be computed for each volume in our proposed datasets.
All saliency maps are computed for the ground truth class, thus avoiding biases due to classification errors.
Following Choe et al. \cite{WSOLRight}, we choose $\delta = 0.5$ when measuring \texttt{MaxBoxAcc} and \texttt{Max3DBoxAcc}, and $\delta \in \{0.3, 0.5, 0.7\}$ when measuring \texttt{MaxBoxAccV2} and \texttt{Max3DBoxAccV2}.

From the results of Table~\ref{tab:main_results}, it is clear that the 3D-specific saliency methods generally perform better than 2D methods. Still, there are cases where the 2D methods show the best localization accuracy. 
%shows that 3D-Grad-CAM++ and 3D-Grad-CAM have comparable or superior performances to the natively 3D Saliency Tubes and Respond-CAM across all metrics. 
We discover, however, that the localization performance of all methods is much lower for 3D CNNs than the state-of-the-art for 2D CNNs on 2D data. Indeed, as reported by~\cite{WSOLRight}, popular saliency maps for 2D images achieve \texttt{MaxBoxAcc} of up to $0.78$ on the CUB benchmark, and  \texttt{PxAP} (equivalent to \texttt{VxAP}) of up to $0.63$ on the OpenImages benchmark. While 2D and 3D benchmarks are not directly comparable, the large discrepancy between~\cite{WSOLRight} and our results in  Table~\ref{tab:main_results} highlights a worrying limitation of all saliency methods when it comes to explaining 3D CNNs. 
On the one hand, this could be due to the amplified localization errors caused by the increased dimensionality. Indeed, localization errors of the same magnitude have higher impact on $IoU$ in 3D space than in 2D space, resulting in lower \texttt{BoxAcc} (\texttt{2D}, \texttt{3D}, and \texttt{V2}). 
On the other hand, the performance discrepancy can also be seen for WSSS metrics \texttt{VxAP} and \texttt{MaxF1}, which are not affected by the increased dimensionality. This is also evident in Figure~\ref{fig:teaser}, where the evaluated saliency methods fail to precisely localize the object in the image, despite the model's correct prediction.
In light of these observations, we hypothesize that there is room for advancement in 3D saliency methods.
Moreover, the consistently superior performance of Respond-CAM and Saliency Tubes suggests that specialized 3D saliency methods hold promise as a potential solution to better explain 3D CNNs.

% This suggests that there is room for improvement in 3D saliency methods.
% Furthermore, the higher performance of Respond-CAM and Saliency Tubes across the board indicates that 3D-specific saliency methods may be a promising solution.

% On the other hand, as depicted in Figure \ref{fig:teaser}, the evaluated saliency methods often fail to precisely localize the object in the image, despite the model's correct prediction.
% This is also demonstrated by the low \texttt{VxAP} and \texttt{MaxF1}, which show that the saliency maps do not sharply focus on the object of interest.

Lastly, the low 2D WSOL metric scores compared to 3D WSOL metrics (Table~\ref{tab:main_results}) highlight the limitations of 2D metrics in assessing the saliency maps at a volume level. This issue becomes particularly apparent when dealing with objects of irregular shape. Indeed, object regions composed of few voxels in different slices may result in substantial 3D GT bounding boxes, even though the object covers a small area in each slice. In such cases, the GT bounding box for each slice may not reflect the region actually covered by the object in the slice, and thus 2D WSOL metrics may be unfairly penalized.

\begin{table}
\caption{Results of the Mass Concentration Sanity Check.}\vspace{-4pt}
\label{tab:cammass}
\centering
\begin{adjustbox}{max width=0.45\textwidth}

    \begin{tabular}{l|lc}
\multicolumn{1}{c|}{\multirow{1}{*}{\makecell[c]{
%Dataset \\ $<$classes$>$ \\ (test accuracy)
}}} & \multicolumn{1}{c}{\multirow{1}{*}{Saliency Method}} & \multicolumn{1}{c}{\makecell[c]{Mass Concentration}} \\
\hline 
\multirow{6}{*}{\rotatebox[origin=c]{0}{\makecell[c]{shapenet-pairs \\ Classes: airplane/bench \\ Test accuracy: $0.967$}}} & & \\[-5pt]
 & Grad-CAM~\cite{Grad-CAM} & $0.752$ \\
 & Grad-CAM++~\cite{Grad-CAM++} & $0.727$ \\
 & HiResCAM~\cite{HiResCAM} & $0.713$ \\
 & Respond-CAM~\cite{Respond-CAM} & $\mathbf{0.793}$ \\
 & SaliencyTubes~\cite{saliency_tubes} & $0.744$ \\[-5pt]
  & & \\
\hline 

    % \begin{tabular}{lll}
    % \multirow{2}{*}{Dataset (accuracy)} & \multirow{2}{*}{saliency method} & \multirow{2}{*}{Mass Concentration} \\
    %  &  &  \\
    % \multirow{5}{*}{shapenet-pairs airplane/bench (0.967)} & GradCAM & 0.752 \\
    %  & GradCAM++ & 0.727 \\
    %  & HiResCAM & 0.713 \\
    %  & Respond-CAM & 0.793 \\
    %  & SaliencyTubes & 0.744
    \end{tabular}
\end{adjustbox}
\vspace{-8pt}
\end{table}

We also test all saliency methods on the ``\texttt{shapenet-pairs}" dataset using the \texttt{MC} metric, and report the results in Table~\ref{tab:cammass}. We find that for all methods, roughly $75\%$ of the saliency mass is correctly located in the half belonging to $\mathbf{x}_C$ to be classified. This indicates that a good portion ($\approx 25\%$) of the saliency map activation mass is located in a completely irrelevant region of the classification volume, and further corroborates our claim that there is margin for improving saliency maps for 3D CNNs. 
% Additional figures supporting our claims are available in the supplementary materials.
Additional figures supporting our claims are available at the project's repository.

\section{Conclusion and Future Works}
\label{sec:conclusion}
% \vspace{-3pt}

We have proposed SE3D: the first framework to rigorously evaluate 3D saliency methods.
We believe that our benchmarks are an important stepping stone towards explainability of 3D CNNs and better solutions to WSOL and WSSS tasks on 3D data. Our results show that current saliency methods are not viable solutions for either task, but that future 3D-specific saliency methods could bridge this gap.

The limitation of saliency maps to explain 3D CNNs is worrying, especially since these models are being used in critical fields. 
As a matter of fact, the worst localization performance of the tested methods was for the medical imaging dataset \texttt{brats-halves}.
Hence, we encourage the research community to develop more accurate and trustworthy explanation methods.

In future works, we intend to address these limitations and develop better saliency methods for 3D CNNs. Lastly, we plan to extend our framework to other saliency methods and datasets, including those of videos, with a particular focus on critical settings such as medical imaging.

%For what strictly concerns our framework, we plan on extending our evaluation to a larger set of datasets and architectures, to provide a more robust assessment. Another research direction consists in extending our metrics to include measures of \textit{insertion} and \textit{deletion}, as introduced in \cite{RISE}. Future developments also include performing user studies to assess the perceived accuracy of 3D saliency methods and investigate further whether these are significantly less accurate than their 2D counterparts.

\vfill\pagebreak

\bibliographystyle{IEEEbib}
\bibliography{ICIP/bibliography}

\begin{thebibliography}{10}

\bibitem{DeepLesion}
K.~Yan, X.~Wang, L.~Lu, and R.~M. Summers,
\newblock ``Deeplesion: Automated deep mining, categorization and detection of
  significant radiology image findings using large-scale clinical lesion
  annotations,'' 2017.

\bibitem{AutonomousDrivingVoxelNet}
Y.~Zhou and O.~Tuzel,
\newblock ``Voxelnet: End-to-end learning for point cloud based 3d object
  detection,''
\newblock in {\em Proceedings of the IEEE conference on computer vision and
  pattern recognition}, 2018, pp. 4490--4499.

\bibitem{WSOLRight}
J.~Choe, S.~J. Oh, S.~Lee, S.~Chun, et~al.,
\newblock ``Evaluating weakly supervised object localization methods right,''
\newblock in {\em CVPR}, 2020, pp. 3133--3142.

\bibitem{shapenet}
A.~X. Chang, T.~Funkhouser, L.~Guibas, P.~Hanrahan, et~al.,
\newblock ``Shapenet: An information-rich 3d model repository,''
\newblock {\em arXiv preprint arXiv:1512.03012}, 2015.

\bibitem{scannet}
A.~Dai, A.~X. Chang, M.~Savva, M.~Halber, et~al.,
\newblock ``Scannet: Richly-annotated 3d reconstructions of indoor scenes,''
\newblock in {\em Proceedings of the IEEE conference on computer vision and
  pattern recognition}, 2017, pp. 5828--5839.

\bibitem{BraTS1}
B.~H. Menze, A.~Jakab, S.~Bauer, J.~Kalpathy-Cramer, et~al.,
\newblock ``The multimodal brain tumor image segmentation benchmark (brats),''
\newblock {\em IEEE TMI}, vol. 34, no. 10, pp. 1993--2024, 2015.

\bibitem{BraTS2}
S.~Bakas, H.~Akbari, A.~Sotiras, M.~Bilello, et~al.,
\newblock ``Advancing the cancer genome atlas glioma mri collections with
  expert segmentation labels and radiomic features,''
\newblock {\em Scientific Data}, vol. 4, no. 1, pp. 170117--170117, 2017.

\bibitem{BraTS3}
S.~Bakas, M.~Reyes, A.~Jakab, S.~Bauer, et~al.,
\newblock ``Identifying the best machine learning algorithms for brain tumor
  segmentation, progression assessment, and overall survival prediction in the
  brats challenge,'' 2019.

\bibitem{saliency_tubes}
A.~Stergiou, G.~Kapidis, G.~Kalliatakis, C.~Chrysoulas, et~al.,
\newblock ``Saliency tubes: Visual explanations for spatio-temporal
  convolutions,''
\newblock in {\em {ICIP}}. sep 2019, {IEEE}.

\bibitem{Respond-CAM}
G.~Zhao, B.~Zhou, K.~Wang, R.~Jiang, et~al.,
\newblock ``Respond-cam: Analyzing deep models for 3d imaging data by
  visualizations,''
\newblock in {\em MICCAI}. Springer, 2018, pp. 485--492.

\bibitem{Grad-CAM}
R.~R. Selvaraju, M.~Cogswell, A.~Das, R.~Vedantam, et~al.,
\newblock ``Grad-cam: Visual explanations from deep networks via gradient-based
  localization,''
\newblock in {\em ICCV}, 2017, pp. 618--626.

\bibitem{Grad-CAM++}
A.~Chattopadhay, A.~Sarkar, P.~Howlader, and V.~N. Balasubramanian,
\newblock ``Grad-cam++: Generalized gradient-based visual explanations for deep
  convolutional networks,''
\newblock in {\em WACV}, 2018, pp. 839--847.

\bibitem{HiResCAM}
R.~L. Draelos and L.~Carin,
\newblock ``Use hirescam instead of grad-cam for faithful explanations of
  convolutional neural networks,'' 2020.

\bibitem{CAM}
B.~Zhou, A.~Khosla, A.~Lapedriza, A.~Oliva, et~al.,
\newblock ``Learning deep features for discriminative localization,''
\newblock in {\em CVPR}, Los Alamitos, CA, USA, June 2016, pp. 2921--2929, IEEE
  Computer Society.

\bibitem{RISE}
V.~Petsiuk, A.~Das, and K.~Saenko,
\newblock ``Rise: Randomized input sampling for explanation of black-box
  models,''
\newblock {\em arXiv preprint arXiv:1806.07421}, 2018.

\bibitem{gradcam_medical_imaging}
H.~Panwar, P.~Gupta, M.~K. Siddiqui, R.~Morales-Menendez, et~al.,
\newblock ``A deep learning and grad-cam based color visualization approach for
  fast detection of covid-19 cases using chest x-ray and ct-scan images,''
\newblock {\em Chaos, Solitons \& Fractals}, vol. 140, pp. 110190, 2020.

\bibitem{cams_medical_imaging}
M.~R. Karim, T.~D{\"o}hmen, M.~Cochez, O.~Beyan, et~al.,
\newblock ``Deepcovidexplainer: Explainable covid-19 diagnosis from chest x-ray
  images,''
\newblock in {\em BIBM}, 2020, pp. 1034--1037.

\bibitem{3d_gradcam1}
S.~Ghadai, A.~Balu, A.~Krishnamurthy, and S.~Sarkar,
\newblock ``Learning and visualizing localized geometric features using 3d-cnn:
  An application to manufacturability analysis of drilled holes,'' 2018.

\bibitem{3d_gradcam2}
C.~Yang, A.~Rangarajan, and S.~Ranka,
\newblock ``Visual explanations from deep 3d convolutional neural networks for
  alzheimer's disease classification,'' 2018.

\bibitem{chakraborty2020detection}
S.~Chakraborty, S.~Aich, and H.-C. Kim,
\newblock ``Detection of parkinson’s disease from 3t t1 weighted mri scans
  using 3d convolutional neural network,''
\newblock {\em Diagnostics}, vol. 10, no. 6, pp. 402, 2020.

\bibitem{gu2022xc}
S.~Gu, V.~Abdelzad, and K.~Czarnecki,
\newblock ``Xc: Exploring quantitative use cases for explanations in 3d object
  detection,''
\newblock {\em arXiv preprint arXiv:2210.11590}, 2022.

\bibitem{saliency_maps_for_segmentation_models}
H.~Saleem, A.~R. Shahid, and B.~Raza,
\newblock ``Visual interpretability in 3d brain tumor segmentation network,''
\newblock {\em Computers in Biology and Medicine}, vol. 133, pp. 104410, 2021.

\bibitem{3d_occlusion}
T.~Uchiyama, N.~Sogi, K.~Niinuma, and K.~Fukui,
\newblock ``Visually explaining 3d-cnn predictions for video classification
  with an adaptive occlusion sensitivity analysis,'' 2022.

\bibitem{gp_unet}
F.~Dubost, H.~Adams, P.~Yilmaz, G.~Bortsova, et~al.,
\newblock ``Weakly supervised object detection with 2d and 3d regression neural
  networks,''
\newblock {\em Medical Image Analysis}, vol. 65, pp. 101767, 2020.

\bibitem{Saliency_Checks}
J.~Adebayo, J.~Gilmer, M.~Muelly, I.~Goodfellow, et~al.,
\newblock ``Sanity checks for saliency maps,'' 2018.

\bibitem{CUB2011}
C.~Wah, S.~Branson, P.~Welinder, P.~Perona, et~al.,
\newblock ``The caltech-ucsd birds-200-2011 dataset,''
\newblock Tech. {R}ep. CNS-TR-2011-001, California Institute of Technology,
  2011.

\bibitem{cleverhans}
S.~Lapuschkin, S.~W{\"a}ldchen, A.~Binder, G.~Montavon, et~al.,
\newblock ``Unmasking clever hans predictors and assessing what machines really
  learn,''
\newblock {\em Nature communications}, vol. 10, no. 1, pp. 1096, 2019.

\bibitem{PV}
L.~Giulivi, M.~Carman, G.~Boracchi, et~al.,
\newblock ``Perception visualization: Seeing through the eyes of a dnn,''
\newblock in {\em Proceedings of BMVC 2021}, pp. 1--13. {BMVA} Press, 2021.

\bibitem{KiTS21}
N.~Heller, F.~Isensee, D.~Trofimova, R.~Tejpaul, et~al.,
\newblock ``The kits21 challenge: Automatic segmentation of kidneys, renal
  tumors, and renal cysts in corticomedullary-phase ct,'' 2023.

\bibitem{medical_decathlon}
M.~Antonelli, A.~Reinke, S.~Bakas, K.~Farahani, et~al.,
\newblock ``The medical segmentation decathlon,''
\newblock {\em Nature Communications}, vol. 13, no. 1, jul 2022.

\bibitem{kits68}
D.~Müller and F.~Kramer,
\newblock ``Miscnn: a framework for medical image segmentation with
  convolutional neural networks and deep learning,''
\newblock {\em BMC Medical Imaging}, vol. 21, 2021.

\bibitem{lungs69}
F.~Isensee, P.~F. Jaeger, S.~A.~A. Kohl, J.~Petersen, et~al.,
\newblock ``nnu-net: a self-configuring method for deep learning-based
  biomedical image segmentation,''
\newblock {\em Nature Methods}, vol. 18, no. 2, pp. 203--211, Feb 2021.

\bibitem{kits85}
F.~Isensee and K.~H. Maier-Hein,
\newblock ``An attempt at beating the 3d u-net,'' 2019.

\bibitem{3d_cnn_model_architecture}
H.~Zunair, A.~Rahman, N.~Mohammed, and J.~P. Cohen,
\newblock ``Uniformizing techniques to process ct scans with 3d cnns for
  tuberculosis prediction,'' 2020.

\end{thebibliography}

\end{document}

\ninept
\maketitle
\noindent
The code and models are distributed publicly {\color{blue} \href{https://github.com/Nexer8/SE3D}{here}}, including a thorough usage guide.
\section{Code}
We publicly release the code for the SE3D framework {\color{blue} \href{https://github.com/Nexer8/SE3D}{here}}. We provide implementations for the extensions to 2D CNN saliency method metrics and our proposed 3D metrics, and enable extensions to custom metrics via straightforward interfaces.
To compute saliency maps, we extend the {\color{blue} \href{https://github.com/jacobgil/pytorch-grad-cam}{pytorch-grad-cam}} framework, which already provides \texttt{PyTorch} implementations for Grad-CAM, Grad-CAM++, and HiResCAM that work on 2D data. We implement and integrate in the common pytorch-grad-cam framework the two 3D-native saliency methods: Respond-CAM and Saliency Tubes, and extend the existing ones to support 3D data. In a similar fashion, more saliency methods can be added to SE3D.

\section{Datasets}
We propose benchmarks on ShapeNet, ScanNet, and BraTS datasets. The datasets are available through the following links:
\begin{itemize}
    \item {\color{blue}\href{https://shapenet.org/}{ShapeNet}} ({\color{blue}\href{http://shapenet.cs.stanford.edu/shapenet/obj-zip/ShapeNetCore.v1.binvox.zip}{voxelized version}})
    \item {\color{blue}\href{http://www.scan-net.org/}{ScanNet}}
    \item {\color{blue}\href{https://www.med.upenn.edu/cbica/brats2020/data.html}{BraTS}}
\end{itemize}
\noindent
We now give more details on the proposed data processing procedures for each dataset, following Section 3.1.

\subsection{ShapeNet}
We use the ``ShapeNetCore" version of the dataset, consisting of $51\,300$ 3D models from one of 51 {\color{blue}\href{https://gist.github.com/tejaskhot/15ae62827d6e43b91a4b0c5c850c168e}{classes}}. All models are stored both in \texttt{mtl} and \texttt{obj} formats. 
A voxelized version is also provided in the original dataset, however, the shape for these volumes is varying. 
Volumes with standardized $32\times32\times32$ shape can instead be obtained from {\color{blue}\href{http://shapenet.cs.stanford.edu/shapenet/obj-zip/ShapeNetCore.v1.binvox.zip}{a different source}}. The latter is the dataset version used for our benchmarks.
Each voxel in this version of the dataset is binary, and is $1$ if the 3D model intersects the voxel, and $0$ otherwise.

For \texttt{shapenet-binary}, we load all \texttt{binvox} files from class folders \texttt{["03001627", "04379243"]}, corresponding to the two randomly chosen classes $\lambda_1$=\texttt{chair}, $\lambda_2$=\texttt{table}, and assign to each sample a binary label, such that $label(\mathbf{x})$ is $0$ if the sample belongs to class \texttt{chair} and is $1$ if it belongs to class \texttt{table}. The segmentation mask corresponds to the object itself, thus, we copy the binary 3D volume as segmentation.

For \texttt{shapenet-pairs}, we load all \texttt{binvox} files from class folders \texttt{["02691156", "02828884"]}, corresponding to $\lambda_1$=\texttt{airplane}, $\lambda_2$=\texttt{bench}, and all other \texttt{binvox} files belonging to $\Lambda \setminus \{\lambda_1, \lambda_2\}$. For each loaded sample in $\{\lambda_1, \lambda_2\}$, we select one random sample in $\Lambda \setminus \{\lambda_1, \lambda_2\}$ with repetition, and one random integer $[0,1]$ indicating the concatenation order. Each resulting sample pair $\mathbf{x}$ composed of $\mathbf{x}_C \in \lambda_1, \lambda_2, \mathbf{x}_N \in \Lambda \setminus \{\lambda_1, \lambda_2\}$ has shape $64\times32\times32$, and binary label $0$ if $\mathbf{x}_C$ belongs to \texttt{airplane} and $1$ if $\mathbf{x}_C$ belongs to \texttt{bench}. The segmentation mask is zero in the $32\times32\times32$ region corresponding to $\mathbf{x}_N$, and is equal to the volume $\mathbf{x}_C$ in its respective region.

The framework is flexible to the choice of $\{\lambda_1, \lambda_2\}$.

\subsection{ScanNet}
\begin{figure}[ht]
    \centering
    \resizebox{0.45\textwidth}{!}{
        \includegraphics{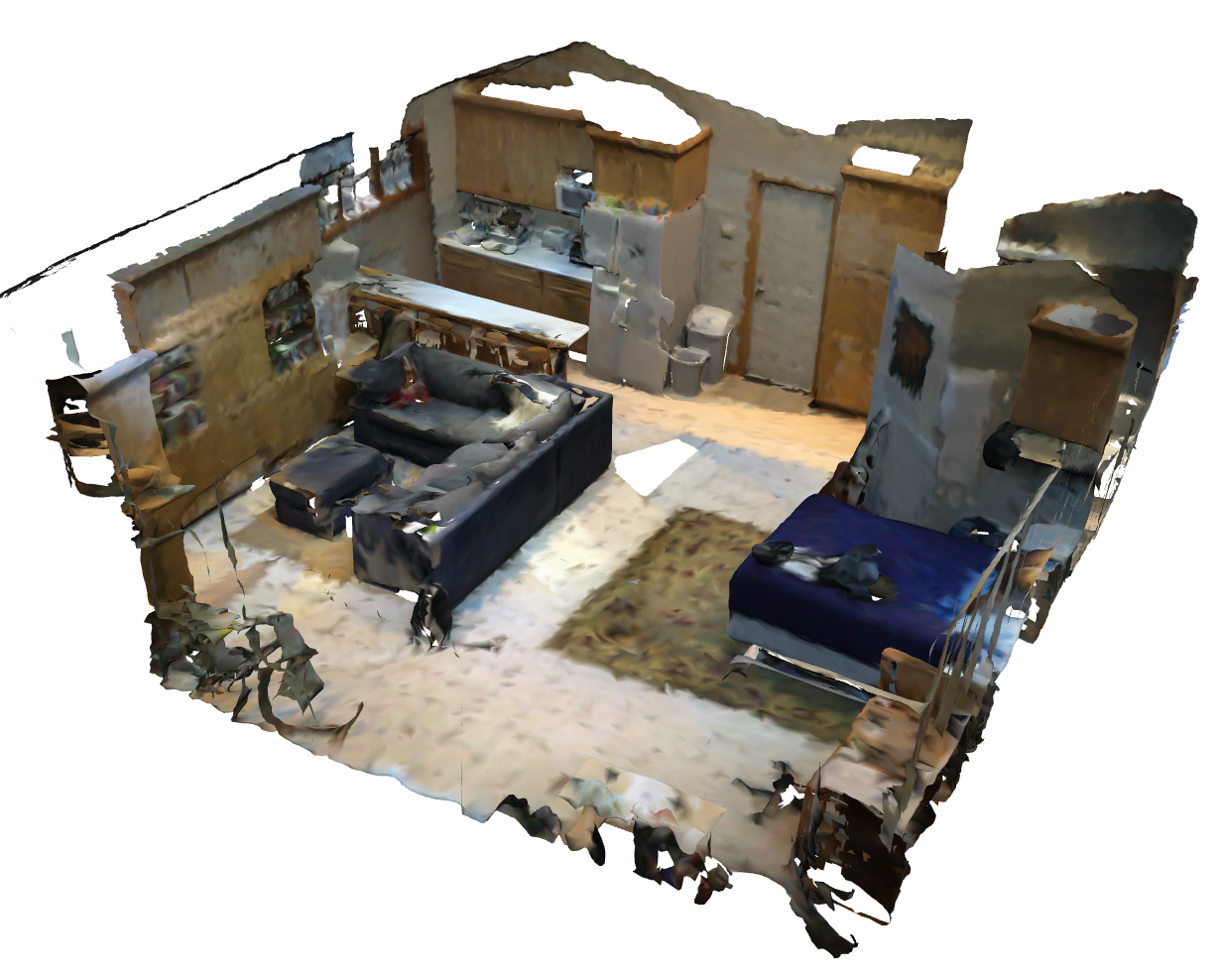}
    }
    \caption{Example scene from ScanNet}
    \label{fig:scannet-scene}
\end{figure}

ScanNet is a dataset of richly-annotated 3D reconstructions of indoor scenes. For our framework, we only require the first scan for each scene, and only the information contained in the \hyphtexttt{['.aggregation.json',} \hyphtexttt{'\_vh\_clean.aggregation.json',} \hyphtexttt{'\_vh\_clean\_2.0.010000.segs.json',} \hyphtexttt{'\_vh\_clean\_2. labels.ply']} files, which contain the meshes and the instance labels.
Each sample in ScanNet is a 3D mesh of an indoor scene (Figure~\ref{fig:scannet-scene}), where groups of points are semantically annotated. Additionally, groups of points are assigned to object instances.

For each scene, we select all instances of objects of classes $\lambda_1$=\texttt{chair}, $\lambda_2$=\texttt{table}. We select the same classes as per ShapeNet to enable interoperability of the models.
For each object instance, we filter the sample's points and only select the points belonging to the object instance, for \texttt{scannet-isolated}, or all the points in the bounding box that surrounds the object instance, for \texttt{scannet-crop}. The filtered meshes are then rescaled to fit in a unit cube, and the latter is voxelized using a $32\times32\times32$ grid, such that the voxel is $1$ if there is at least one mesh point at the corresponding grid location, and $0$ otherwise. The sample's label is $0$ if the object instance is of class \texttt{chair}, and $1$ if the object is of class \texttt{table}. The segmentation mask is the voxelized version of the filtered mesh, as per \texttt{scannet-isolated}, thus excluding environment points.

\subsection{BraTS}
BraTS is a multimodal brain tumor segmentation dataset. Volumetric CT scans are provided as NIfTI files, which are readable using the {\color{blue} \href{https://nipy.org/nibabel/}{NiBabel}} python library.
The scans are provided in the form of 4D volumes, with the fourth dimension representing the different MRI modalities.
The modalities are:
\begin{itemize}
    \item native,
    \item post-contrast T1-weighted,
    \item T2-weighted,
    \item T2 Fluid Attenuated Inversion Recovery.
\end{itemize}
The ground truth segmentation masks are provided in the form of 3D volumes, with each voxel containing a label from the
following set:
\begin{itemize}
    \item $0$: background,
    \item $1$: necrotic and non-enhancing tumor core,
    \item $2$: peritumoral edema,
    \item $4$: GD-enhancing tumor.
\end{itemize}
The preprocessing steps are implemented in the \hyphtexttt{prepare\_data set.py} script. The script takes as input the path to the training folder of the BraTS 2020 dataset, and creates a folder containing the preprocessed dataset.

Each volume in the dataset is preprocessed and split at the \textit{corpus callosum} to obtain the \texttt{brats-halves} dataset. First, each volume is rotated by $-90^{\circ}$ around the $x$ and $y$ axes to align the volumes with the standard radiological view. Volumes are then normalized to the range $[0, 1]$ and then converted to the range $[0, 255]$ to reduce the memory footprint of the dataset. The segmentation masks are converted to contain only two labels: background ($0$) and tumor ($1$). This is done by merging the labels $1$, $2$ and $4$ into a single label. The intuition behind this is that the tumor is the only region of interest for the saliency methods, and the different labels are not relevant for the evaluation. Lastly, both volumes and segmentation masks are split into 2 halves along the \textit{corpus callosum} plane. For each resulting half, the label is set to $1$ if at least one voxel is labeled as tumor, and $0$ otherwise.

The BraTS dataset contains CT scans of anonymous subjects, and is publicly available. As such, no ethical concerns arise on the usage of this dataset for benchmarking purposes.

\section{Models and Training}
\begin{figure}[ht]
    \centering
    \resizebox{0.45\textwidth}{!}{
        \includegraphics{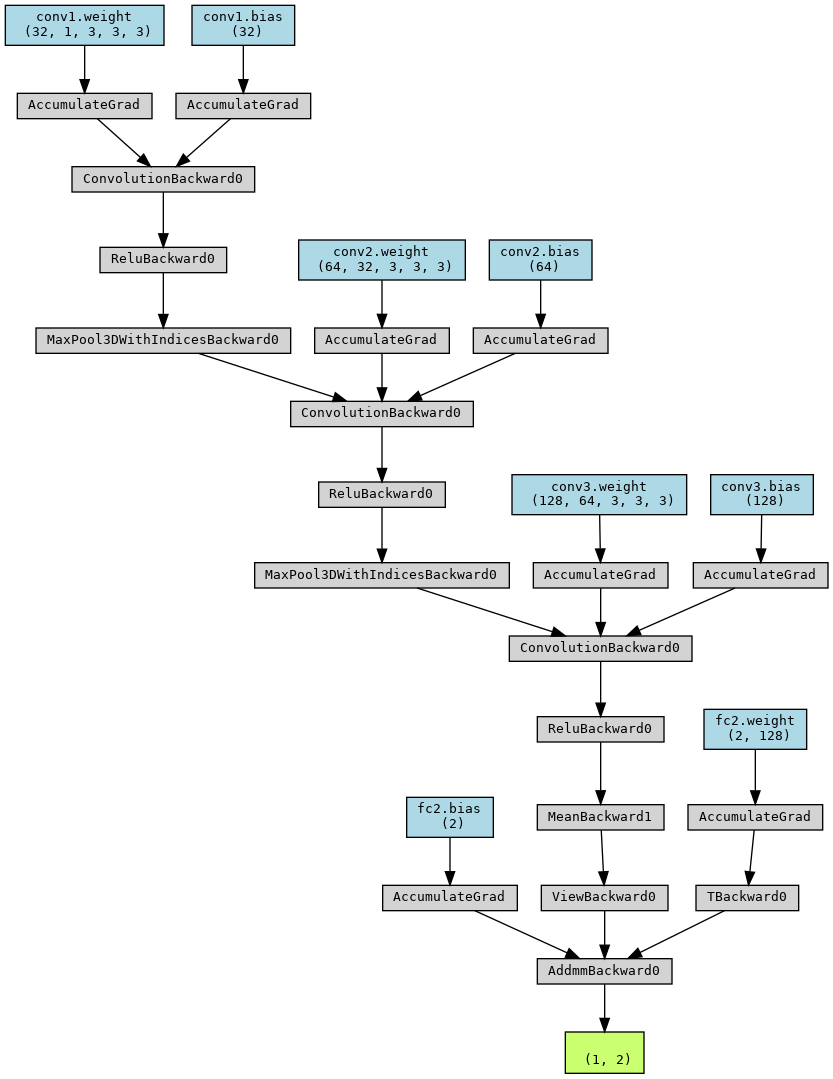}
    }
    \caption{Model architecture}
    \label{fig:model-arch}
\end{figure}

For all datasets, we use the same architecture, composed of three ReLU-activated 3D convolutional layers interleaved with $2\times2$ pooling layers, and a GAP+FC head (Firgure~\ref{fig:model-arch}). While a more sophisticated architecture could enable slightly better performance, our focus is to evaluate saliency maps, and not the models. Thus, we keep the model simple to avoid biases that might influence the computation of saliency maps, while still achieving good results. 
For ShapeNet and ScanNet, it is enough to train the model for $1$ epoch to saturate performance, while for BraTS we train for $20$ epochs. We use the AdamW optimizer on a constant schedule with $lr=0.0001$, $wd=0.01$. We use a class-weighted loss and perform augmentation by random affine transformations using the {\color{blue}\href{https://torchio.readthedocs.io/}{torchio}} library.

All model checkpoints are publicly available {\color{blue} \href{https://github.com/Nexer8/SE3D}{here}}.

\section{Saliency Map Visualization}
\begin{figure}[t]
    \centering
    \renewcommand\tabcolsep{0pt}
    \renewcommand\theadfont{\large}
    \begin{adjustbox}{max width=0.65\textwidth}
        %\hspace{0.00\textwidth}
        \begin{tabularx}{\textwidth}{cc}
            \makecell[c]{\setlength{\fboxsep}{10pt}\setlength{\fboxrule}{0.0pt}\fbox{\includegraphics[width=170pt,align=c]{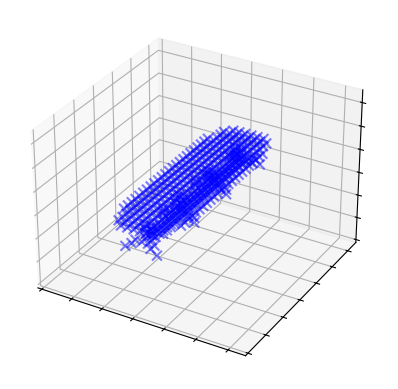}} \\ \Large Sample} & 
            \makecell[c]{\setlength{\fboxsep}{10pt}\setlength{\fboxrule}{0.0pt}\fbox{\includegraphics[width=170pt,align=c]{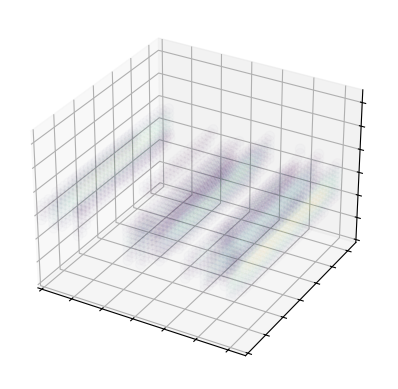}} \\ \Large Grad-CAM} \\
            \makecell[c]{\setlength{\fboxsep}{10pt}\setlength{\fboxrule}{0.0pt}\fbox{\includegraphics[width=170pt,align=c]{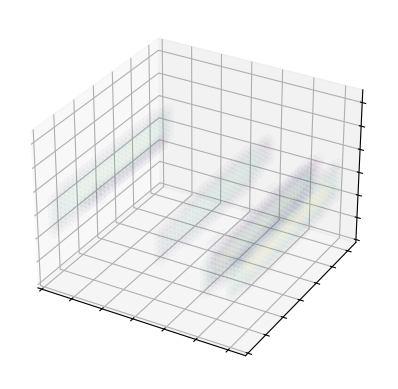}} \\ \Large Grad-CAM++} & 
            \makecell[c]{\setlength{\fboxsep}{10pt}\setlength{\fboxrule}{0.0pt}\fbox{\includegraphics[width=170pt,align=c]{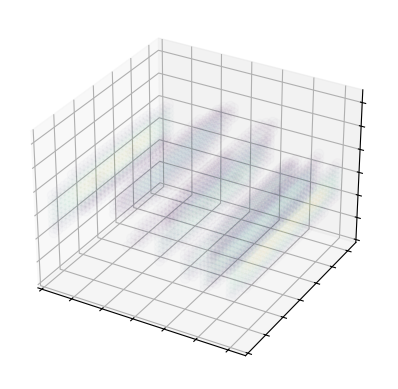}} \\ \Large HiResCAM} \\
            \makecell[c]{\setlength{\fboxsep}{10pt}\setlength{\fboxrule}{0.0pt}\fbox{\includegraphics[width=170pt,align=c]{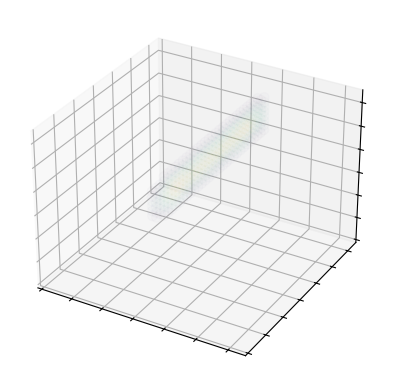}} \\ \Large Respond-CAM} & 
            \makecell[c]{\setlength{\fboxsep}{10pt}\setlength{\fboxrule}{0.0pt}\fbox{\includegraphics[width=170pt,align=c]{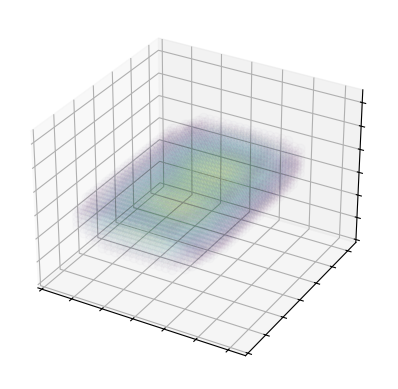}} \\ \Large Saliency Tubes} \\
        \end{tabularx}
    \end{adjustbox}
    \vspace{0pt}
    \caption{Example saliency maps for \texttt{shapenet-binary}, class \texttt{table}. Saliency maps are thresholded at $\tau=0.3$ to avoid clutter in visualization. It is clear how most saliency methods are scattered and focus on irrelevant portions of the image. Saliency Tubes is the only method that focuses on the correct region, confirming our findings in Table 1.}
    \label{fig:ex_binary_1}
    \vspace{-5pt}
\end{figure}

\begin{figure}[t]
    \centering
    \renewcommand\tabcolsep{0pt}
    \renewcommand\theadfont{\large}
    \begin{adjustbox}{max width=0.65\textwidth}
        %\hspace{0.00\textwidth}
        \begin{tabularx}{\textwidth}{cc}
            \makecell[c]{\setlength{\fboxsep}{10pt}\setlength{\fboxrule}{0.0pt}\fbox{\includegraphics[width=170pt,align=c]{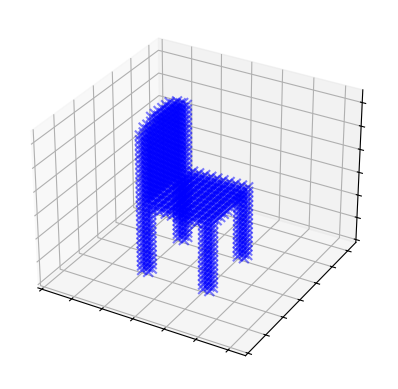}} \\ \Large Sample} & 
            \makecell[c]{\setlength{\fboxsep}{10pt}\setlength{\fboxrule}{0.0pt}\fbox{\includegraphics[width=170pt,align=c]{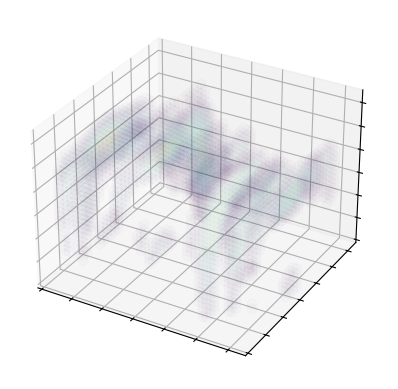}} \\ \Large Grad-CAM} \\
            \makecell[c]{\setlength{\fboxsep}{10pt}\setlength{\fboxrule}{0.0pt}\fbox{\includegraphics[width=170pt,align=c]{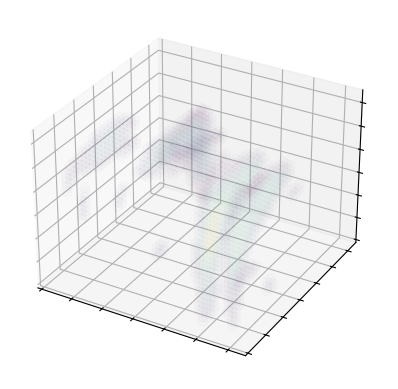}} \\ \Large Grad-CAM++} & 
            \makecell[c]{\setlength{\fboxsep}{10pt}\setlength{\fboxrule}{0.0pt}\fbox{\includegraphics[width=170pt,align=c]{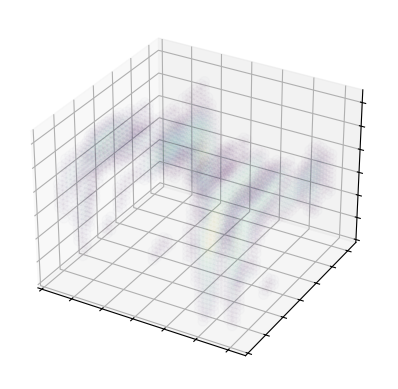}} \\ \Large HiResCAM} \\
            \makecell[c]{\setlength{\fboxsep}{10pt}\setlength{\fboxrule}{0.0pt}\fbox{\includegraphics[width=170pt,align=c]{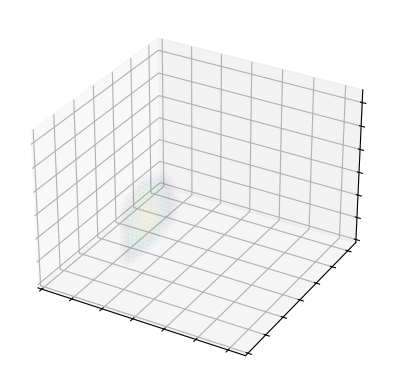}} \\ \Large Respond-CAM} & 
            \makecell[c]{\setlength{\fboxsep}{10pt}\setlength{\fboxrule}{0.0pt}\fbox{\includegraphics[width=170pt,align=c]{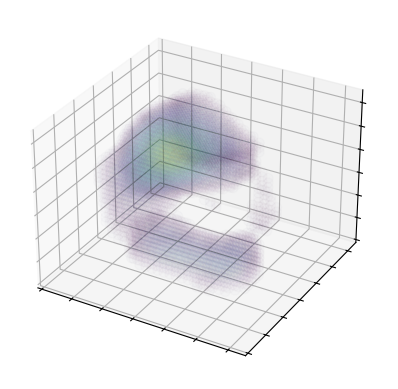}} \\ \Large Saliency Tubes} \\
        \end{tabularx}
    \end{adjustbox}
    \vspace{0pt}
    \caption{Example saliency maps for \texttt{shapenet-binary}, class \texttt{chair}. Saliency maps are thresholded at $\tau=0.3$ to avoid clutter in visualization. Also in this case, the saliency maps for all methods except Saliency Tubes do not localize well the object in the volume.\\ \\}
    \label{fig:ex_binary_2}
    \vspace{-5pt}
\end{figure}

We provide additional figures displaying how the saliency maps fail to localize the object.
Our first example (Figure~\ref{fig:ex_binary_1}) displays saliency maps for a sample in \texttt{shapenet-pairs} of class \texttt{table}, computed for all tested methods. Most saliency maps do not localize well the bench in the volume.
Similar results can be seen also in the second example (Figure~\ref{fig:ex_binary_2}) for a sample of class \texttt{table}.
We highlight an emerging pattern, namely, extensions of 2D methods to 3D (Grad-CAM, Grad-CAM++, HiResCAM) are very dispersed throughout the volume, while Respond-CAM displays very little if any activation on most of the volume. Saliency Tubes, instead, shows the best performance in both cases.

\begin{figure}[t]
    \centering
    \renewcommand\tabcolsep{0pt}
    \renewcommand\theadfont{\large}
    \begin{adjustbox}{max width=0.65\textwidth}
        %\hspace{0.00\textwidth}
        \begin{tabularx}{\textwidth}{cc}
            \makecell[c]{\setlength{\fboxsep}{10pt}\setlength{\fboxrule}{0.0pt}\fbox{\includegraphics[width=170pt,align=c]{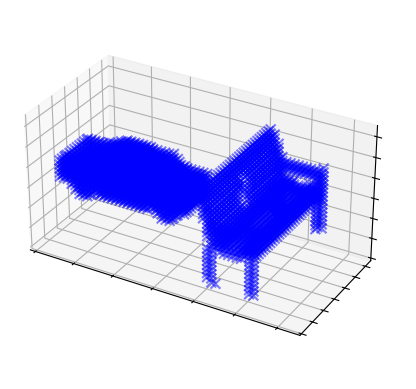}} \\ \Large Sample} & 
            \makecell[c]{\setlength{\fboxsep}{10pt}\setlength{\fboxrule}{0.0pt}\fbox{\includegraphics[width=170pt,align=c]{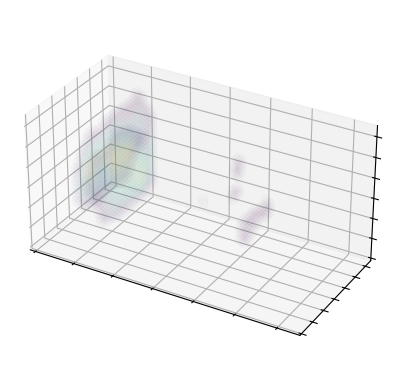}} \\ \Large Grad-CAM} \\
            \makecell[c]{\setlength{\fboxsep}{10pt}\setlength{\fboxrule}{0.0pt}\fbox{\includegraphics[width=170pt,align=c]{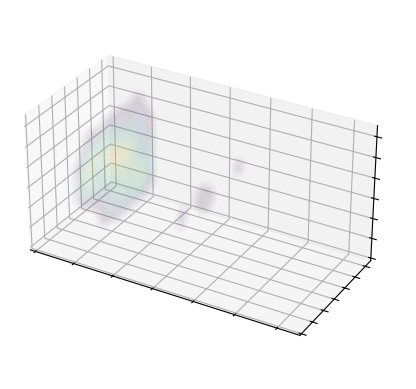}} \\ \Large Grad-CAM++} & 
            \makecell[c]{\setlength{\fboxsep}{10pt}\setlength{\fboxrule}{0.0pt}\fbox{\includegraphics[width=170pt,align=c]{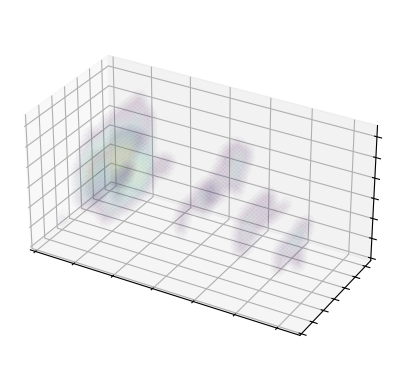}} \\ \Large HiResCAM} \\
            \makecell[c]{\setlength{\fboxsep}{10pt}\setlength{\fboxrule}{0.0pt}\fbox{\includegraphics[width=170pt,align=c]{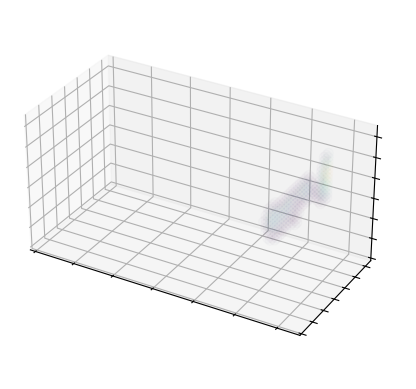}} \\ \Large Respond-CAM} & 
            \makecell[c]{\setlength{\fboxsep}{10pt}\setlength{\fboxrule}{0.0pt}\fbox{\includegraphics[width=170pt,align=c]{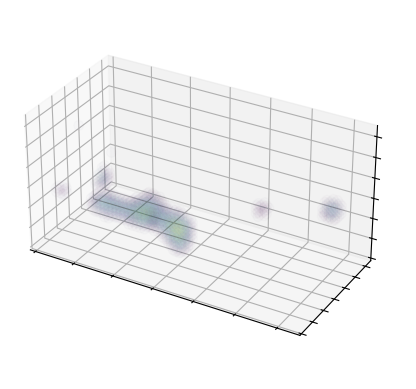}} \\ \Large Saliency Tubes} \\
        \end{tabularx}
    \end{adjustbox}
    \vspace{0pt}
    \caption{Example saliency maps for \texttt{shapenet-pairs}, class \texttt{bench} (on the right). Saliency maps are thresholded at $\tau=0.3$ to avoid clutter in visualization. Most of the methods focus on the wrong side of the volume.}
    \label{fig:ex_pairs_1}
    \vspace{-5pt}
\end{figure}

For \texttt{shapenet-pairs}, we see similar results. We show an example in Figure~\ref{fig:ex_pairs_1}, where we display saliency maps for a sample where $\mathbf{x}_C$ is of class \texttt{bench} and is concatenated as the second element (to the right in the volume). Most of the saliency maps incorrectly focus on the left side of the volume, except for Respond-CAM. This confirms our results in Table 2 of the paper, where Respond-CAM showed the best \texttt{MC} performance.

\begin{figure}[t]
    \centering
    \renewcommand\tabcolsep{0pt}
    \renewcommand\theadfont{\large}
    \begin{adjustbox}{max width=0.65\textwidth}
        %\hspace{0.00\textwidth}
        \begin{tabularx}{\textwidth}{cc}
            \makecell[c]{\setlength{\fboxsep}{10pt}\setlength{\fboxrule}{0.0pt}\fbox{\includegraphics[width=170pt,align=c]{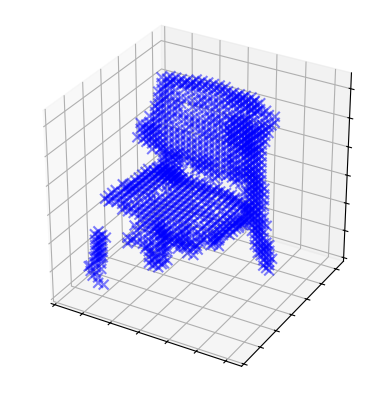}} \\ \Large Sample} & 
            \makecell[c]{\setlength{\fboxsep}{10pt}\setlength{\fboxrule}{0.0pt}\fbox{\includegraphics[width=170pt,align=c]{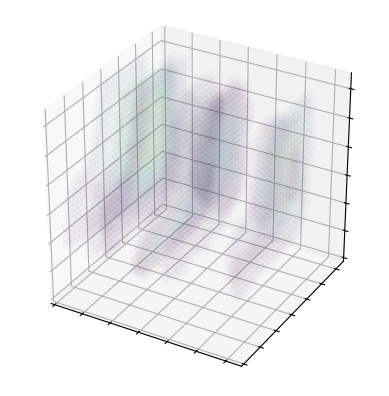}} \\ \Large Grad-CAM} \\
            \makecell[c]{\setlength{\fboxsep}{10pt}\setlength{\fboxrule}{0.0pt}\fbox{\includegraphics[width=170pt,align=c]{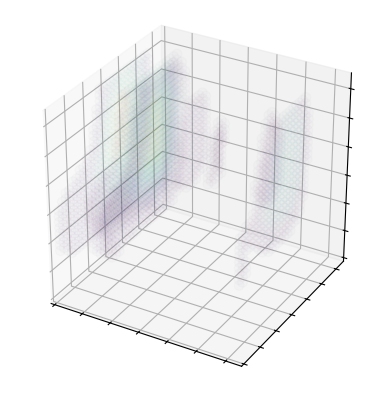}} \\ \Large Grad-CAM++} & 
            \makecell[c]{\setlength{\fboxsep}{10pt}\setlength{\fboxrule}{0.0pt}\fbox{\includegraphics[width=170pt,align=c]{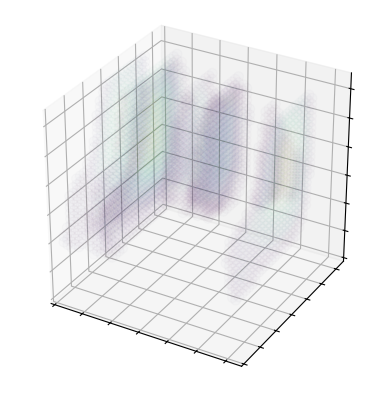}} \\ \Large HiResCAM} \\
            \makecell[c]{\setlength{\fboxsep}{10pt}\setlength{\fboxrule}{0.0pt}\fbox{\includegraphics[width=170pt,align=c]{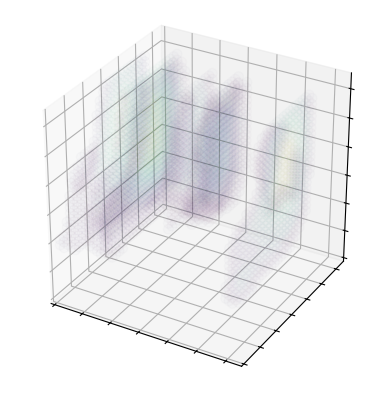}} \\ \Large Respond-CAM} & 
            \makecell[c]{\setlength{\fboxsep}{10pt}\setlength{\fboxrule}{0.0pt}\fbox{\includegraphics[width=170pt,align=c]{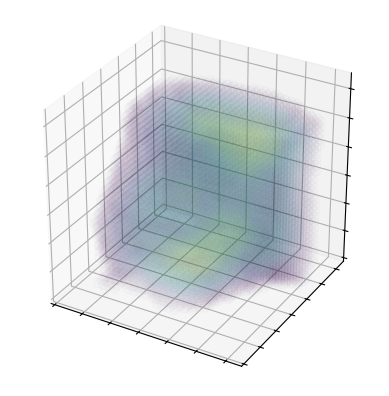}} \\ \Large Saliency Tubes} \\
        \end{tabularx}
    \end{adjustbox}
    \vspace{0pt}
    \caption{Example saliency maps for \texttt{scannet-isolated}, class \texttt{chair}. Saliency maps are thresholded at $\tau=0.3$ to avoid clutter in visualization. The object is large enough that even very dispersed saliency maps cover the relevant regions. Despite this, we still see how Saliency Tubes provides a much more refined localization.}
    \label{fig:ex_isolated_1}
    \vspace{-5pt}
\end{figure}

The pattern continues also for \texttt{scannet-isolated}, as shown in Figure~\ref{fig:ex_isolated_1}. In the example, the object is very large, thus, all class activation maps, even if dispersive, focus on relevant regions. Nonetheless, Saliency Tubes still displays the best localization accuracy.

% \vfill\pagebreak

% % References should be produced using the bibtex program from suitable
% % BiBTeX files (here: strings, refs, manuals). The IEEEbib.bst bibliography
% % style file from IEEE produces unsorted bibliography list.
% % -------------------------------------------------------------------------
% \bibliographystyle{IEEEbib}
% \bibliography{ICIP/bibliography}